%% file: main.tex
\documentclass[10pt,twocolumn,letterpaper]{article}

\usepackage[pagenumbers]{iccv} 

\definecolor{iccvblue}{rgb}{0.21,0.49,0.74}
\usepackage[pagebackref,breaklinks,colorlinks,allcolors=iccvblue]{hyperref}
\usepackage{tcolorbox} 
\tcbuselibrary{breakable}
\usepackage{multirow}
\usepackage[table]{xcolor} 
\usepackage{wrapfig}
\usepackage{fontawesome}
\usepackage{placeins}

\input{preamble}

\newcommand{\tablestyle}[2]{\setlength{\tabcolsep}{#1}\renewcommand{\arraystretch}{#2}\centering\footnotesize}
\newlength\savewidth\newcommand\shline{\noalign{\global\savewidth\arrayrulewidth
		\global\arrayrulewidth .8pt}\hline\noalign{\global\arrayrulewidth\savewidth}}		
\newcommand\scline[1]{\noalign{\global\savewidth\arrayrulewidth
		\global\arrayrulewidth .8pt}\cline{#1}\noalign{\global\arrayrulewidth\savewidth}}

\usepackage[normalem]{ulem}
\usepackage[capitalize]{cleveref}
\crefname{section}{Sec.}{Secs.}
\Crefname{section}{Section}{Sections}
\Crefname{table}{Table}{Tables}
\crefname{table}{Tab.}{Tabs.}
\Crefname{figure}{Figure}{Figures}
\crefname{figure}{Fig.}{Figs.}
\crefname{appendix}{Appx.}{Appxs.}
\newcommand{\worldwideweb}{\raisebox{-1.5pt}{\includegraphics[height=1.05em]{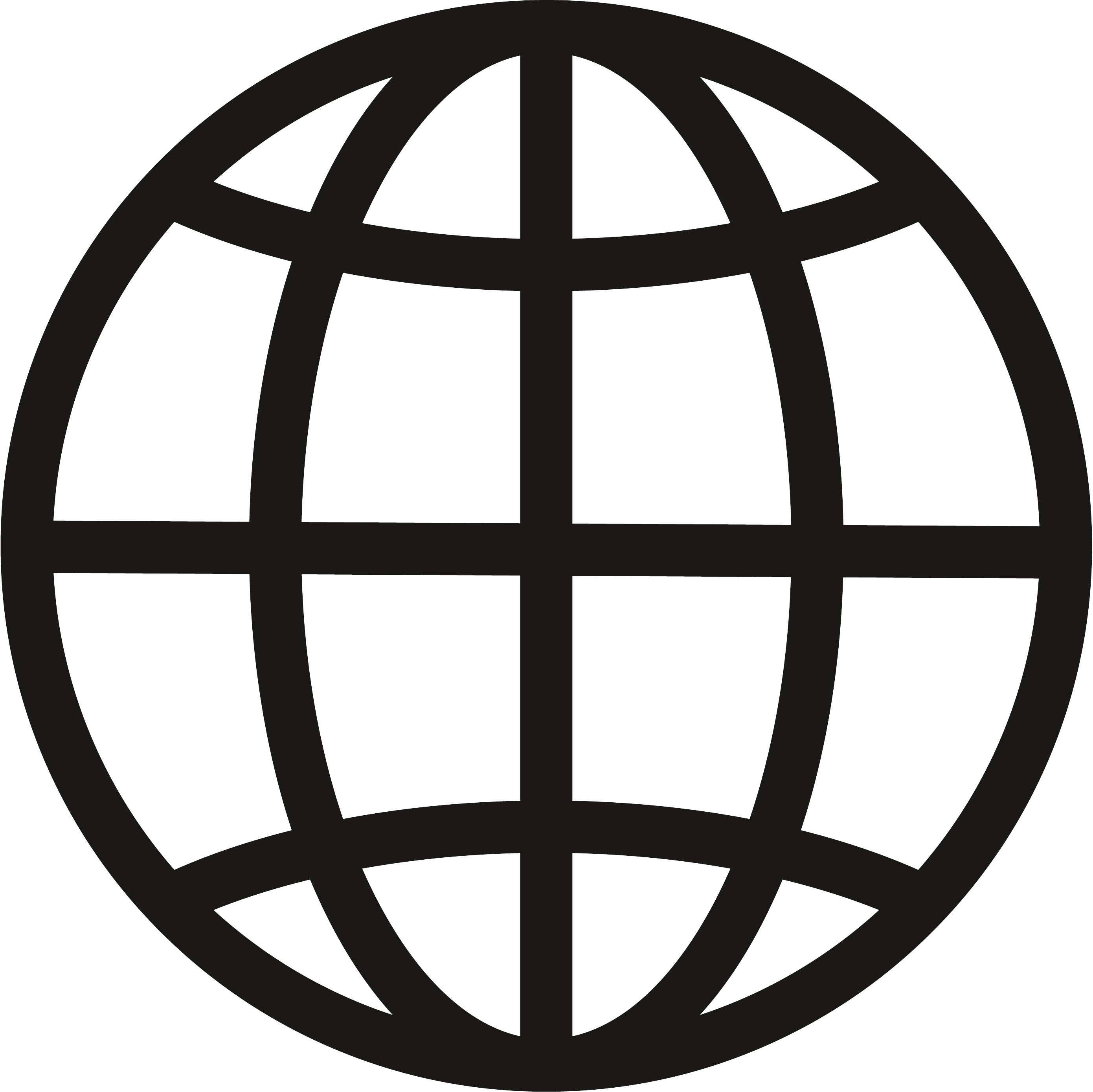}}\xspace}
\newcommand{\github}{\raisebox{-1.5pt}{\includegraphics[height=1.05em]{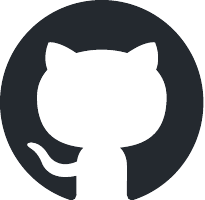}}\xspace}
\newcommand{\huggingface}{\raisebox{-1.5pt}{\includegraphics[height=1.05em]{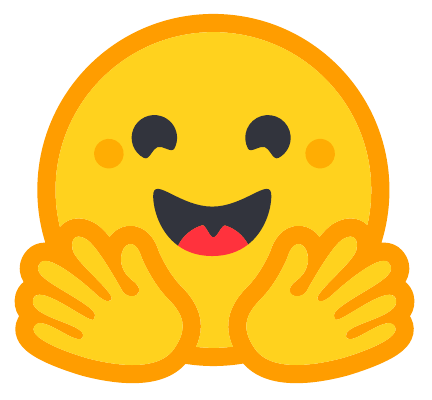}}\xspace}

\title{LEGO-Puzzles: How Good Are MLLMs at Multi-Step Spatial Reasoning?}

\author{
\bf Kexian Tang$^{1,3}$\footnotemark[1] \qquad Junyao Gao$^{2,3}$\footnotemark[1] \qquad Yanhong Zeng$^{3}$\footnotemark[2] \qquad Haodong Duan$^{3}$\footnotemark[2] \\
\bf Yanan Sun$^{3}$ \qquad Zhening Xing$^{3}$ \qquad Wenran Liu$^{3}$ \qquad Kai Chen$^{3}$ \qquad Kaifeng Lyu$^{1}$\footnotemark[3] \\
Tsinghua University$^1$ \qquad
Tongji University$^2$ \qquad
Shanghai AI Laboratory$^3$ \\
{\tt\small tangkx25@mails.tsinghua.edu.cn, klyu@mail.tsinghua.edu.cn} \\
\\
{\worldwideweb \href{https://tangkexian.github.io/LEGO-Puzzles/}{\text{Project Page}}} \quad
{\github \href{https://github.com/Tangkexian/LEGO-Puzzles}{\text{Code}}} \quad
{\huggingface \href{https://huggingface.co/datasets/KexianTang/LEGO-Puzzles}{\text{LEGO-Puzzles}}}
}

\begin{document}
\maketitle

\renewcommand{\thefootnote}{\fnsymbol{footnote}}
\footnotetext[1]{Equal contribution.}
\footnotetext[2]{Project Leads.}
\footnotetext[3]{Corresponding Author.}

\makeatletter
\renewcommand{\thefootnote}{\arabic{footnote}}
\makeatother

\input{0_abstract}

\input{1_intro}

\input{2_related_work}

\input{3_method}

\input{4_experiment}

\input{5_generation}

\input{6_discussion}

\input{7_conclusion}

\FloatBarrier
{
    \small
    \bibliographystyle{ieeenat_fullname}
    \bibliography{main}
}

\appendix
\input{8_appendix}


\end{document}

%% file: preamble.tex
\newcommand{\imagetoken}[1]{\texttt{\char`<image #1\char`>}}
\newcommand{\promptnewline}{\texttt{\char`\\n}}

\newtcolorbox{promptbox}[1][]{
  breakable,
  colback=gray!4,
  colframe=black!45,
  boxrule=0.5pt,
  arc=1mm,
  left=1.5mm,
  right=1.5mm,
  top=1mm,
  bottom=1mm,
  fonttitle=\bfseries,
  fontupper=\small,
  title={#1}
}

\newtcolorbox{taskrubric}[1]{
  breakable,
  colback=gray!3,
  colframe=black!35,
  colbacktitle=gray!18,
  coltitle=black,
  boxrule=0.45pt,
  arc=1mm,
  left=1.5mm,
  right=1.5mm,
  top=1mm,
  bottom=1mm,
  before skip=5pt,
  after skip=5pt,
  fonttitle=\bfseries,
  title={#1}
}

%% file: 0_abstract.tex
\begin{abstract}
\label{sec:abs}

Many real-world applications of spatial intelligence, such as robotic control, autonomous driving, and automated assembly, require spatial reasoning across multiple sequential steps.
However, the extent to which current Multimodal Large Language Models~(MLLMs) possess this capability remains largely unexplored.
Inspired by LEGO construction, a recreational activity that critically relies on multi-step spatial reasoning, we introduce \textbf{LEGO-Puzzles}: a benchmark designed to systematically evaluate the spatial reasoning capabilities of MLLMs from basic spatial understanding to multi-step planning.
LEGO-Puzzles contains two task sets. The \textbf{Elementary} set covers $11$ visual question-answering~(VQA) tasks with $1,100$ carefully curated samples to test elementary spatial reasoning skills that are cruical for LEGO assembly. The \textbf{Planning} set directly requires the model to generate a step-by-step plan for assembling a target LEGO structure, where the tasks are organized into subsets with different planning horizons ranging up to $8$.
Our evaluation of 29 state-of-the-art MLLMs shows that even the strongest models struggle with elementary reasoning tasks in LEGO construction, falling at least $20\%$ behind human performance. 
The planning accuracy also quickly drops to $0\%$ as the number of planning steps increases, whereas our human participants solve all the tasks perfectly.
Switching the output format from multiple choice to image generation degrades model performance even further, leading to zero accuracy even for planning $3$ steps.
Overall, LEGO-Puzzles reveals critical limitations in current MLLMs' spatial reasoning capabilities and highlights the need for substantial advances.


\end{abstract}

%% file: 1_intro.tex
\section{Introduction}\label{sec:intro}

Multimodal Large Language Models (MLLMs) have made remarkable progress and have shown promising capabilities in spatial intelligence~\cite{gardner2011frames}.
The most elementary capability of spatial intelligence is the ability to perceive and understand spatial relationships among 3D objects from a static scene, which we refer to as \textbf{spatial understanding} in this work.
This includes many Visual Question Answering (VQA) tasks, such as identifying object attributes (e.g., shape, color, size), recognizing spatial relations (e.g., left/right, above/below), and counting objects that satisfy certain conditions~\cite{johnson2017clevr,li2023super,wang20233d,liao2024reasoning}.

Many other tasks not only require the spatial understanding of the given scene, but also critically rely on the ability to imagine the consequences of hypothetical actions applied to the scene. That is, one has to sequentially apply the spatial understanding to two or more real or imaginary scenes to solve these tasks, a capability which we refer to as \textbf{sequential spatial reasoning} in this work.
Notably, a series of benchmarks have been introduced to evaluate the ability of imagining the scene after changing the camera perspective~\cite{shiri2024empirical,ma20243dsrbench,stogiannidis2025mind,li2025viewspatial}.

However, only a few benchmarks~\cite{yang2024thinking,su2024actplan,ma2025phyblock} evaluate the sequential spatial reasoning capability of MLLMs beyond one step of action, even though many real-world applications require planning for multiple steps of actions.
For example, to assemble a target structure, a robotic arm may perceive the current and target states and plan a long sequence of actions to execute.
In home maintenance scenarios, users may send a photo of a broken item to a chatbot for help, and the chatbot needs to plan a sequence of actions for the user to follow step by step.
Providing a systematic evaluation of the spatial intelligence of MLLMs from single-step to multiple-step reasoning is hence the focus of this work.

\paragraph{LEGO-Puzzles.} In this work, we take inspiration from a common recreational activity, \textit{LEGO construction},
to assess the spatial intelligence of MLLMs.
Constructing LEGO models naturally requires reasoning over sequential assembly steps, making it a convenient testbed for evaluating spatial reasoning abilities.
Moreover, LEGO construction has been widely used in cognitive science as a reliable indicator of spatial intelligence. Studies have shown that LEGO construction is strongly associated with core spatial skills such as \emph{mental rotation} and \emph{visuo-spatial working memory}, as well as with mathematical performance~\cite{mcdougal2024assessing,mcdougal2023associations,jirout2015building}. 

\begin{figure*}[t]
\centering
\includegraphics[width=1\linewidth]{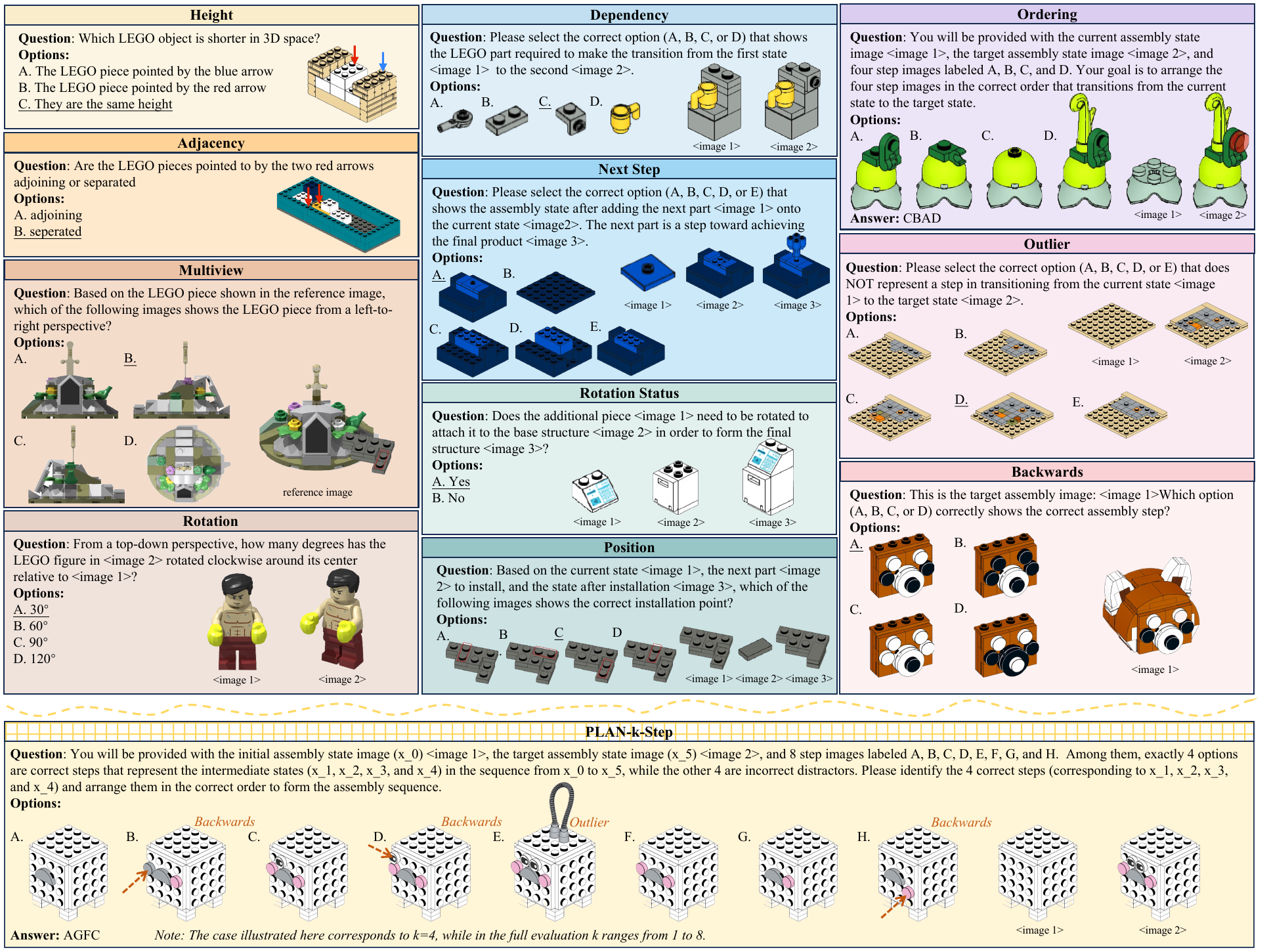}
\caption{
\small{\textbf{Task examples of LEGO-Puzzles}. The examples above the wavy line correspond to the \emph{Elementary Set}, with columns (from left to right) representing Spatial Understanding, Single-Step Sequential Reasoning, and Multi-Step Sequential Reasoning. The examples below the wavy line correspond to the \emph{Planning Set}. \textit{Note: The questions are slightly simplified for clarity and brevity.}}}
\label{fig:all}
\vspace{-5mm}
\end{figure*}

We introduce \textbf{LEGO-Puzzles}, a benchmark designed to systematically evaluate the spatial reasoning capabilities of MLLMs with progressively increasing levels of spatial and sequential complexity, ranging from basic spatial understanding to multi-step planning.
Based on open-source LEGO projects with step-by-step assembly instructions, we curate a diverse set of tasks organized into two sets, the \textbf{Elementary} set and the \textbf{Planning} set (see \cref{fig:all}).
\vspace{2mm}

\begin{itemize}[label=\textbullet,
  leftmargin=1.8em,
  labelindent=0.2em,
  labelsep=0.5em, 
  itemsep=0ex,topsep=0.3ex]
    \item The \textbf{Elementary} set contains a diverse collection of over 1,100 carefully curated visual question-answering (VQA) pairs spanning 11 distinct tasks, grouped into three levels. The first level assesses MLLMs' \textbf{basic spatial understanding} capabilities, 
including recognition of height relationships, rotation angles, etc. The second and third levels test \textbf{single-step} and \textbf{multi-step sequential reasoning} abilities of MLLMs based on LEGO assembly sequences to examine models' sequential reasoning ability. 
\vspace{2mm}
    \item The \textbf{Planning} set is especially designed to assess whether a model's capability of multi-step spatial reasoning is strong enough to generate an explicit multi-step plan for assembling a target LEGO structure.
    The task is organized into subsets with different numbers of planning steps $k$, which ranges up to $8$.
    While the primary evaluation is conducted in a multiple-choice format, the task also evaluates MLLMs' ability to generate images of the planned intermediate states in multiple turns, mimicking real-world scenarios where a chatbot guides users through a sequence of actions using visual instructions.
\end{itemize}
\vspace{-3mm}

\paragraph{Evaluation Results.}
We conduct comprehensive evaluations of 29 state-of-the-art MLLMs, including proprietary models such as GPT-5 and Gemini-2.5-Pro, as well as leading open-source alternatives~\cite{chen2024far,wang2024qwen2,wu2024deepseek}.
The evaluation results reveal significant limitations in the spatial reasoning capabilities of MLLMs.
    Even the strongest models lag behind humans by at least $20\%$ on the tasks in the elementary set. Among open-source models, only a few achieve performance notably above random guessing.
This human-model gap becomes more pronounced in the planning tasks, where the multiple-choice accuracy quickly drops below $50\%$ once the number of steps exceeds $3$, and eventually approaches $0\%$ as the number of steps further increases.
In contrast, human participants achieve perfect accuracy on all these planning tasks.
When switching the output format from multiple choice to image generation, MLLMs can usually generate images with the correct appearance, but fail to place LEGO pieces in the correct positions. As a result, the accuracy drops to zero even for planning $3$ steps.

\paragraph{Our Contributions.}
Our main contributions are as follows:
\begin{itemize}[label=\textbullet,
  leftmargin=1.8em,
  labelindent=0.2em,
  labelsep=0.5em, 
  itemsep=0ex,topsep=0.3ex]


\item \textbf{Comprehensive evaluation framework.}
Our benchmark includes a diverse set of tasks with progressively increasing spatial and sequential complexity, ranging from basic spatial understanding to multi-step planning.


\item \textbf{Natural but challenging planning tasks.}
Our planning tasks are naturally constructed from open-source human-designed LEGO projects but can pose significant challenges for frontier MLLMs to solve.

\item \textbf{Evaluation of multi-turn image generation.}
Beyond VQA evaluation, our planning tasks also evaluate the ability of MLLMs to generate planned intermediate states as images in multiple turns. All the frontier MLLMs fail completely even for $3$ planning steps.

\end{itemize}


%% file: 2_related_work.tex
\section{Related Work}
\paragraph{General Multi-Modal Evaluation Benchmarks.} Recent years have seen significant advancements in MLLMs, accompanied by a surge in benchmark datasets evaluating their visual understanding. MME \cite{Fu2023MMEAC} provides a systematic evaluation of 14 image-centric tasks, revealing persistent challenges such as object hallucination and spatial reasoning failures. MMBench \cite{liu2024mmbench} introduces a bilingual multiple-choice format for fine-grained multimodal assessment. Moving beyond static images, SEED-Bench \cite{li2023seed} evaluates generative comprehension spanning both image and video reasoning. For expert-level reasoning, MMMU \cite{yue2024mmmu} presents a discipline-specific benchmark across 183 subtopics, revealing substantial knowledge gaps even in leading MLLMs. Overall, these benchmarks reveal that while MLLMs have made progress, they still struggle with spatial understanding and long-horizon reasoning, presenting clear directions for future research.

\vspace{-2mm}
\paragraph{Visual Spatial Reasoning Evaluation Benchmarks.}
Multimodal large language models (MLLMs) have made notable progress on vision-and-language tasks, yet they still struggle with 3D spatial reasoning. 
Beyond general benchmarks, several more specific datasets have been proposed to evaluate the spatial reasoning abilities of MLLMs. 
CLEVR~\cite{johnson2017clevr} evaluates diverse visual reasoning skills with minimal biases and fine-grained annotations.
More recently,
3DSRBench~\cite{ma20243dsrbench} designs 12 tasks assessing spatial understanding from perspectives such as relative height, location, and orientation.
SpatialViz-Bench~\cite{wang2025spatialviz} provides a cognitively grounded benchmark of 12 tasks targeting four spatial skills: rotation, folding, penetration, and animation.
Despite their significance, most of these efforts primarily focus on \emph{single-image} spatial reasoning.

\vspace{-2mm}
\paragraph{Multi-Step Spatial Reasoning Evaluation Benchmarks.}
Several benchmarks incorporate sequential aspects of spatial reasoning.
VSI-Bench~\cite{yang2024thinking} introduces a video-based benchmark evaluate MLLMs’ visual-spatial intelligence from sequential observations.
ActPlan-1K \cite{su2024actplan} focuses on procedural planning for household activities given images and task descriptions.
PhyBlock~\cite{ma2025phyblock} proposes a progressive benchmark for evaluating VLMs on physical understanding and planning. 
Although it includes tasks requiring multi-step planning, the planning tasks in our benchmark are substantially more complex and challenging. 
In PhyBlock, the best-performing models achieve over 40\% accuracy, whereas in our benchmark even the strongest models fail completely. 
Even when partial correctness is considered, model performance in our benchmark exceeds random guessing by less than 20\%. 
Moreover, the tasks in PhyBlock are conducted in relatively simplified or highly abstract environments—only 8 shapes, 5 colors, and at most 22 pieces per scene. 
In contrast, LEGO-Puzzles uses diverse, open-source LEGO files spanning: very small builds ($<10$ pieces), medium structures, and large, complex assemblies ($>100$ pieces), with more than 30 colors and over 100 different brick shapes.
OrigamiSpace~\cite{xu2025origamispace} evaluates the multi-step spatial reasoning ability of MLLMs and their capacity to handle mathematical constraints through origami tasks. Similarly, GamiBench~\cite{spencer2025gamibench} studies spatial reasoning and 2D-to-3D planning using origami-inspired folding tasks. However, the tasks in these benchmarks remain easier than those in our setting, where the best-performing models achieve over 50\% accuracy.
Despite their contributions, all the above benchmarks do not evaluate the multi-turn image generation ability of MLLMs, which is crucial for real-world applications such as robotic manipulation and automated assembly.
Extending beyond prior benchmarks, our work establishes a testbed that directly evaluates MLLMs' fine-grained multi-step spatial reasoning abilities, offering a more comprehensive assessment of long-horizon spatial reasoning and full-stage planning.

%% file: 3_method.tex
\section{LEGO-Puzzles}

\label{sec:met}
\label{sec:task}
In this section, we introduce LEGO-Puzzles, a benchmark designed to evaluate the multi-step spatial reasoning capability of MLLMs through a progressive and comprehensive evaluation framework.
The benchmark contains two sets: (1) the \textbf{Elementary set}, which includes a diverse set of LEGO-related VQA tasks spanning spatial understanding, single-step and multi-step sequential reasoning~(\cref{sec:task-Elementary Set}); (2) the \textbf{Planning set}, which is specifically designed to evaluate whether a model's reasoning capability is strong enough to generate an explicit multi-step plan for assembling a target LEGO structure~(\cref{sec:task-Planning Set}).
Our data curation process is described in \cref{sec:data}.

\subsection{Elementary Set}

\label{sec:task-Elementary Set}
The Elementary Set is designed to test fundamental saptial reasoning skills that are crucial for LEGO assembly.
To enable a more comprehensive and progressively structured evaluation, we define three levels of tasks. This framework is grounded in insights from cognitive psychology and human developmental stages in acquiring spatial intelligence \cite{newcombe2010early,gardner2011frames}.
Using LEGO building as a concrete and intuitive example, we observe that humans typically develop spatial reasoning abilities in stages: from basic \textbf{\textit{spatial understanding (Level 1)}}, to reasoning through \textbf{\textit{individual assembly steps (Level 2)}}, and ultimately to reasoning across \textbf{\textit{multiple sequential steps (Level 3)}}. Based on this developmental trajectory, our benchmark is divided into three levels, as illustrated in \cref{fig:all}.

\subsubsection{Level 1: Spatial Understanding.}
This level focuses on the ability to \textit{understand the spatial relationships} between each LEGO piece and how the pieces appear and relate when viewed from different perspectives in 3D space:
(1) \textit{Height:} Distinguish the relative heights of LEGO objects.
(2) \textit{Adjacency:} Determine whether LEGO objects are adjacent or separated.
(3) \textit{Rotation:} Calculate the angle of rotation between a LEGO object and its corresponding rotated version.
(4) \textit{Multiview:} Predict the current LEGO status from different viewpoints.


\subsubsection{Level 2: Single-Step Sequential Reasoning.}
Building upon spatial understanding, this level 
is designed based on single-step actions, mirroring how humans typically progress from spatial perception to carrying out one assembly step at a time during the building process:
(5) \textit{Rotation Status:} Determine whether a LEGO piece requires rotation before installation. In contrast to \textit{Rotation} Task in Level 1, this task focuses on reasoning from an assembly perspective, with finer granularity.
(6) \textit{Position:} Identify the correct assembly position to place the next LEGO piece.
(7) \textit{Next-Step:} Predict the next LEGO status based on the current status and the new pieces. 
(8) \textit{Dependency:} Identify which pieces are necessary to transition from the current to the next assembly stage.

\begin{figure*}[!t]

\centering
\includegraphics[width=\linewidth]{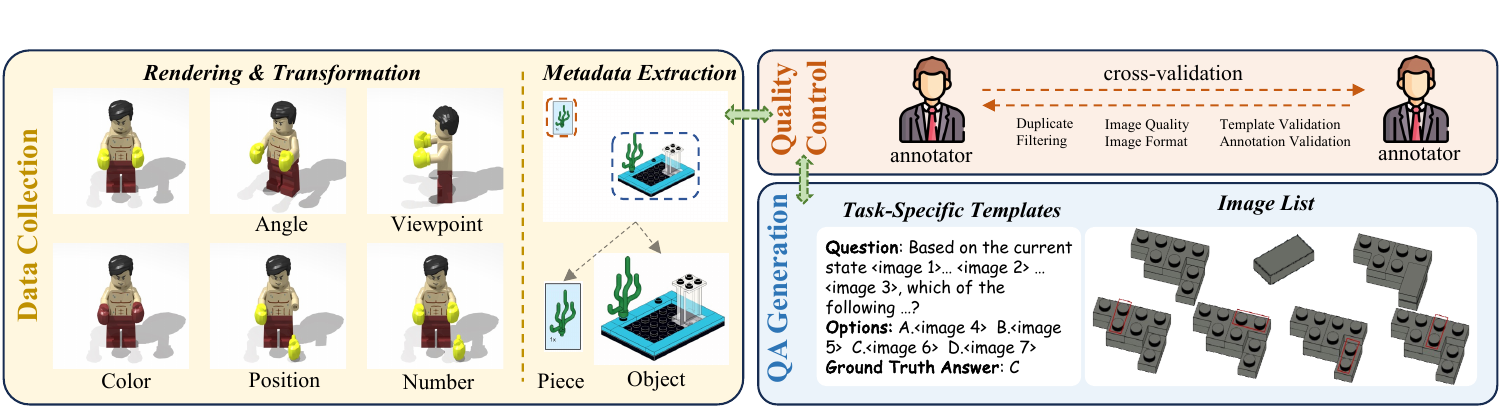}
\caption{\small{\textbf{Data curation pipeline.} Our pipeline comprises: data collection, question-answer generation, and quality control.}}
\label{fig:data}

\end{figure*}
\subsubsection{Level 3: Multi-Step Sequential Reasoning.}
The final level is built upon the single-step reasoning skills from \textit{Level 2} and require planning over multiple sequences (involving up to 7 intermediate stages). It assesses the ability to reason across multiple steps in the assembly process:
(9) \textit{Backwards:} Given the target assembly stage, determine the correct intermediate step in the LEGO build process.
(10) \textit{Ordering:} Determine the correct assembly order of the provided final LEGO images.
(11) \textit{Outlier:} Detect the LEGO status that does not belong to the provided assembly sequence.

\subsubsection{Task Summary.}
The Elementary set consists of over $1,\!100$ visual question-answering (VQA) pairs derived from $407$ LEGO building instructions, encompassing $11$ tasks across spatial understanding, single-step and multi-step sequential reasoning. Each task contains $100$ samples to ensure balanced evaluation across categories. 


\subsection{Planning Set}

\label{sec:task-Planning Set}
To further investigate the long-horizon multi-step sequential reasoning abilities of MLLMs, we construct the Planning Set, which goes beyond elementary spatial reasoning by requiring step-by-step plan. Within this set, we design two tasks: \textbf{Plan-$k$-Step-VQA} and \textbf{Plan-$k$-Step-Generation}, which evaluate the model's ability to generate a multi-step plan in VQA and image generation settings, respectively.

\subsubsection{Task: Plan-$k$-Step-VQA.}
\textit{PLAN-k-Step-VQA} integrates the multi-step tasks \textit{Backwrds}, \textit{Ordering} and \textit{Outlier} from the Elementary set to model the realistic reasoning planning and execution across multiple sequences under noisy conditions.
\textit{PLAN-$k$-Step-VQA} first extends the original \textit{Ordering} setting, which was restricted to a fixed \textit{PLAN-4-Step}, into a scalable sequence length $k$ setting ranging from 1 to 8.
Next, we add an additional $k$ erroneous options from the tasks \textit{Backwrds} and \textit{Outlier} on each $k$-length sequence, resulting in a total of $2k$ options.
These noisy conditions increase the task's complexity and make the examination of MLLMs' multi-step reasoning ability more realistic and challenging.
Specifically, we construct 20 test cases for every sequence length k. Each test case includes the {current LEGO object} ($x_0$), the {target LEGO object} ($x_{k+1}$), the {correct intermediate LEGO objects} ($x_1^{c}, x_2^{c}, \dots, x_k^{c}$), and the {erroneous LEGO objects} ($x_1^{e}, x_2^{e}, \dots, x_k^{e}$), along with the corresponding text instructions. The model is required to select the $k$ correct intermediate steps and order them according to the assembly sequence.

\subsubsection{Task: Plan-$k$-Step-Generation.}
To further investigate whether MLLMs can transfer their sptial reasoning abilities to image generation and perform realistic full-stage planning, we design a more challenging image generation task: \textit{PLAN-$k$-Step-Generation}. It mirrors how humans naturally perform LEGO assembly:
given an initial assembly state and a final target state, an intelligent system should generate the intermediate states step by step to demonstrate how the build is completed. 
Motivated by this idea, we design an interactive, step-by-step generative framework, where the model produces one intermediate state at each turn. Given the initial state $x_0$ and the final target state $x_{k+1}$:

\begin{itemize}[label=\textbullet,
  leftmargin=1.8em,
  labelindent=0.2em,
  labelsep=0.5em, 
  itemsep=0ex,topsep=0.3ex]
\item Turn 1: The model receives $x_0$ and $x_{k+1}$ and is asked to generate the next assembly state $x_1$.
\item Turn 2: The model is then asked to generate $x_2$, conditioned on its previously generated output $x_1$.
\item Turn (t): At each subsequent turn, the model generates $x_{t+1}$ based on the entire history of generated states.
\end{itemize}
This process continues until the model produce $x_k$, yielding a complete generative trajectory. We set $k = 2, 3, 4, 5$. 
We do not evaluate longer horizons because even at $k=3$ the strongest models already fail completely (\cref{sec:generation}).

\subsection{Dataset Curation}

\label{sec:data}

As illustrated in \cref{fig:data}, 
our pipeline consists of three key steps: data collection, question-answer generation, and quality control. This design ensures the scalability, accuracy, and reliability of our data.

\subsubsection{Data Collection.} 
Data collection consists of three stages.
\textbf{\textit{(1). LEGO Project Collection.}}
We collect a diverse set of open-source LEGO source files from the Internet, each containing detailed step-by-step building instructions and part lists. 
To ensure suitable task complexity, we filter for projects with moderate final size. 
Extremely large builds exhibit high structural complexity, and the small visual changes introduced by added pieces make it difficult for models to detect step-wise differences.
Conversely, overly small builds are filtered out for low spatial complexity and insufficient steps for multi-step reasoning.
We also ensure category diversity, covering animals, furniture, vehicles, and more, to increase task variety.
\textbf{\textit{(2). Rendering and Transformation.}}
We render each project into PDF format using the public software Studio\footnote{\url{https://www.bricklink.com/v3/studio/download.page}}, keeping the camera viewpoint fixed across steps to maintain spatial and temporal consistency. 
The tool allows flexible editing of the source files, enabling us to modify part attributes such as type, quantity, color, and position as needed for task construction. 
For example, in the \textit{Rotation} and \textit{Multiview} task, we apply POV-Ray style rendering and adjust lighting to simulate different viewing angles. In the \textit{Backward} task, we introduce deliberate errors in part attributes to generate incorrect assembly states.
\textbf{\textit{(3).Metadata Extraction and Unification.}}
We employ PDF-Extract\footnote{\url{https://github.com/opendatalab/PDF-Extract-Kit}} to extract structured information from the rendered PDFs, including individual LEGO pieces and assembled objects. All visual assets are processed under a unified naming convention and organized for downstream question-answer generation across all defined task types.

\subsubsection{Question-Answer Generation.} 
To support scalable and structured data construction, we manually design task-specific templates tailored to each task. Each example includes an instruction defining the model's role, a question referencing input images using tokens like \imagetoken{x}, and a ground-truth answer.
For example, in the \textit{Position} task, the model is provided with the current assembly state, the part to install, the resulting state, and several candidate placements. 
This standardized template design allows flexibility in the number of input images required by each task, making it suitable for both single-step and multi-step reasoning scenarios. More details are provided in the appendix.

\subsubsection{Quality Control.}
To ensure data quality and reliability, we implement a multi-stage human-in-the-loop review process.
\textit{1)Duplicate filtering.} We perform duplication checks by computing similarity scores across rendered images, identifying visually redundant or recolored samples. These are further reviewed manually, and duplicates are removed to reduce redundancy and improve evaluation quality.
\textit{2)Image Quality and Format Check.} We manually verify all rendered images to ensure consistency in camera perspective, correctness of part attributes (e.g., color, shape, quantity), and adherence to naming conventions. Any files with errors are either corrected or discarded.
\textit{3)Template and annotation validation.}
Each question-answer pair is verified by three trained annotators. Reviewers confirm that the images referenced by tokens (\imagetoken{1}, \imagetoken{2}, \ldots) appear in the correct order. Samples with unresolved disagreements are either revised or removed.

Given the vast number of high-quality open-source LEGO projects available, our pipeline is inherently scalable and can be extended in an semi-automated manner to support larger and more diverse benchmarks in the future.

%% file: 4_experiment.tex
\section{Experiment on the Elementary Set}

\label{sec:ex}
\subsection{Setup}

\paragraph{Benchmark Models.}
We extensively evaluate 29 models, covering a diverse range of architectures, sizes, and training processes. 
For open-source models, we evaluate Qwen3-VL-[8B/32B]-Instruct~\cite{bai2025qwen3}, Qwen2.5-VL-[7B/72B]~\cite{bai2025qwen2}, Qwen2-VL-[7B/72B]~\cite{wang2024qwen2}, InternVL3.5-[8B/38B]~\cite{wang2025internvl3}, InternVL3-[8/78B]~\cite{zhu2025internvl3}, InternVL2.5-[8B/78B]~\cite{chen2024expanding}, MiniCPM-V2.6~\cite{yao2024minicpm}, VILA1.5-13B~\cite{lin2024vila}, Idefics3-8B~\cite{laurenccon2024building}, DeepSeek-VL2-[Tiny/Small]~\cite{wu2024deepseek}, Pixtral-12B~\cite{agrawal2024pixtral}, LLaVA-OneVision-7B~\cite{li2024llava}, and EMU3~\cite{wang2024emu3}. 
For proprietary models, we evaluate GPT-5
~\cite{openai2025gpt5}, GPT-o3~\cite{openai2025gpto3}, GPT-4o
, GPT-4o-mini~\cite{OpenAI2023GPT4TR},  Claude-3.5-Sonnet~\cite{Claude3}, Gemini-2.5-Pro~\cite{comanici2025gemini}, Gemini-2.0-Flash, Gemini-1.5-Flash~\cite{team2023gemini}, and Gemini-1.5-Pro.
All evaluations are conducted in a zero-shot setting for a fair comparison. 

\vspace{-3mm}
\paragraph{Baselines.}
We provide two baselines for comparison:

\begin{itemize}[label=\textbullet,
  leftmargin=1.8em,
  labelindent=0.2em,
  labelsep=0.5em, 
  itemsep=0ex,topsep=0.3ex]
    \item \textit{Random}: the accuracy of random selection, assuming equal probability for all options.
    \item  $\uparrow$ \textit{Random}: the \textit{p-value–based critical value}, which indicates the minimum accuracy required to statistically surpass random guessing at a given significance level ($p = 0.05$).
\end{itemize}

\vspace{-3mm}
\paragraph{Evaluation Metrics.}
For all questions in the Elementary set, we adopt \textit{exact match accuracy} (\%) as the evaluation metric. 
For multiple-choice questions, we follow VLMEvalKit~\cite{duan2024vlmevalkit}, applying rule-based matching as a first step, and resort to LLM-based choice extraction and matching using ChatGPT~\cite{OpenAI2023GPT4TR} when heuristic matching fails. 
For the \textit{Ordering} task, we adopt the same strategy: rule-based extraction of the predicted step sequence from the model’s response, followed by LLM-based extraction and matching when necessary. 
\begin{table*}[!t]
    \centering
    \caption{\small\textbf{Full evaluation results of 29 MLLMs on the Elementary set.} {\colorbox{gray!70}{\small Dark Gray} and \colorbox{gray!30}{\small Light Gray} indicates the best performance for each task among all models and open-source models respectively. We highlight the top 3 models by overall performace  using \colorbox{green}{\small Dark Green}, \colorbox{green!30}{\small Medium Green}, and \colorbox{green!10}{\small Light Green}, respectively.}}
    \vspace{-3mm}
    \resizebox{\textwidth}{!}{%
    \tablestyle{3pt}{1.3}
    \begin{tabular}{lccccccccccc|r}
    \shline
    \multirow{2}{*}{\textbf{Models}} & \multicolumn{4}{c}{\cellcolor{cyan!5}\textbf{Spatial Understanding}} & \multicolumn{4}{c}{\cellcolor{orange!10}\textbf{Single-Step Reasoning}} & \multicolumn{3}{c|}{\cellcolor{brown!15}\textbf{Multi-Step Reasoning}} & \multirow{2}{*}{\textbf{Overall}}\\
    & Height & Adjacency & Rotation & Multiview & Next-Step & Dependency & Rotation Stat. & Position & Backwards & Ordering & Outlier \\
    \shline
    \rowcolor{yellow!15}\multicolumn{13}{l}{\textit{\textbf{Proprietary}}} \\
    GPT-5 & 62.0 & 69.0 & \cellcolor{gray!70}72.0 & \cellcolor{gray!70}67.0 & \cellcolor{gray!70}83.0 & \cellcolor{gray!70}92.0 & 55.0 & 59.0 & \cellcolor{gray!70}73.0 & \cellcolor{gray!70}85.0 & \cellcolor{gray!70}75.0 & \cellcolor{green} 72.0 \\
    GPT-o3 & \cellcolor{gray!70}66.0 & \cellcolor{gray!70}71.0 & 64.0 & 66.0 & \cellcolor{gray!70}83.0 & \cellcolor{gray!70}92.0 & 53.0 & 54.0 & 66.0 & 80.0 & 65.0 & \cellcolor{green!30}69.1 \\
    GPT-4o & 49.0 & 66.0 & 41.0 & 51.0 & 65.0 & 87.0 & 51.0 & 51.0 & 53.0 & 72.0 & 49.0 & 57.7 \\
    GPT-4o-mini & 31.0 & 53.0 & 26.0 & 51.0 & 27.0 & 71.0 & 57.0 & 32.0 & 50.0 & 7.0 & 27.0 & 39.3  \\
    Claude-3.5-Sonnet & 39.0 & 60.0 & 42.0 & 48.0 & 61.0 & 78.0 & 58.0 & 37.0 & 49.0 & 54.0 & 64.0 &  53.6 \\
    Gemini-2.5-Pro & 57.0 & 64.0 & 65.0 & 65.0 & 79.0 & 88.0 & 64.0 & \cellcolor{gray!70}67.0 & 59.0 & 82.0 & 67.0 & \cellcolor{green!10} 68.8 \\
    Gemini-2.0-Flash & 35.0 & 70.0 & 49.0 & 45.0 & 69.0 & 81.0 & 54.0 & 46.0 & 56.0 & 46.0 & 43.0 & 54.0 \\
    Gemini-1.5-Flash & 29.0 & 58.0 & 28.0 & 45.0 & 57.0 & 77.0 & 57.0 & 32.0 & 28.0 & 20.0 & 51.0 & 43.8 \\
    Gemini-1.5-Pro & 35.0 & 58.0 & 38.0 & 56.0 & 59.0 & 84.0 & 61.0 & 39.0 & 35.0 & 44.0 & 59.0 & 51.6 \\
    \shline
    \rowcolor{yellow!15}\multicolumn{13}{l}{\textit{\textbf{Open-source}}} \\
    Qwen3-VL-32B-Instruct & 38.0 & 59.0 & 41.0 & \cellcolor{gray!30}58.0 & \cellcolor{gray!30}62.0 & \cellcolor{gray!30}91.0 & \cellcolor{gray!70}65.0 & \cellcolor{gray!30}58.0 & \cellcolor{gray!70}73.0 &  \cellcolor{gray!30}34.0 & \cellcolor{gray!30}48.0 & \cellcolor{green} 57.0 \\
    Qwen3-VL-8B-Instruct & 35.0 & 67.0 & 30.0 & 42.0 & \cellcolor{gray!30}62.0 & 81.0 & 58.0 & 40.0 & 41.0 &  22.0 & 30.0 & 46.2 \\
    Qwen2.5-VL-72B & 30.0 & 61.0 & 27.0 & 27.0 & 55.0 & 72.0 & 58.0 & 47.0 & 60.0 & 33.0 & 43.0 & \cellcolor{green!10}46.6 \\
    Qwen2.5-VL-7B & 35.0 & 60.0 & 22.0 & 27.0 & 26.0 & 60.0 & 49.0 & 25.0 & 24.0 & 5.0 & 13.0 & 31.5 \\
    Qwen2-VL-72B & 40.0 & 62.0 & 37.0 & 51.0 & 57.0 & 79.0 & 49.0 & 43.0 & 34.0 & 26.0 & 31.0 &  46.3 \\
    Qwen2-VL-7B & 31.0 & 57.0 & 30.0 & 40.0 & 44.0 & 70.0 & 48.0 & 26.0 & 13.0 & 9.0 & 28.0 & 36.0 \\
    InternVL3.5-VL-38B & 42.0 & 51.0 & \cellcolor{gray!30}42.0 & 51.0 & 61.0 & 74.0 & 56.0 & 32.0 & 53.0 &  5.0 & 20.0 & 44.3 \\
    InternVL3.5-VL-8B & 43.0 & 54.0 & 26.0 & 36.0 & 41.0 & 52.0 & 57.0 & 31.0 & 32.0 &  6.0 & 33.0 & 37.4 \\
    InternVL3-VL-78B & \cellcolor{gray!30}49.0 & 65.0 & 37.0 & 54.0 & 60.0 & 80.0 & 61.0 & 34.0 & 25.0 &  20.0 & 40.0 & \cellcolor{green!30}47.7 \\
    InternVL3-VL-8B & 39.0 & 58.0 & 24.0 & 42.0 & 30.0 & 57.0 & 53.0 & 25.0 & 30.0 &  3.0 & 33.0 & 35.8 \\
    InternVL2.5-78B & 41.0 & 62.0 & 32.0 & 47.0 & 60.0 & 79.0 & 58.0 & 32.0 & 40.0 & 15.0 & 37.0 & 45.7 \\
    InternVL2.5-8B & 35.0 & 53.0 & 23.0 & 37.0 & 38.0 & 48.0 & 64.0 & 25.0 & 35.0 & 0.0 & 29.0 & 35.2 \\
    MiniCPM-V2.6 & 26.0 & 56.0 & 22.0 & 44.0 & 34.0 & 50.0 & 51.0 & 29.0 & 23.0 & 0.0 & 19.0 & 32.2 \\
    VILA1.5-13B & 26.0 & 55.0 & 26.0 & 35.0 & 17.0 & 34.0 & 48.0 & 26.0 & 12.0 & 4.0 & 22.0 & 27.7 \\
    Idefics3-8B & 29.0 & 51.0 & 23.0 & 23.0 & 18.0 & 20.0 & 47.0 & 30.0 & 24.0 & 4.0 & 24.0 & 26.6 \\
    DeepSeek-VL2-Small & 31.0 & 52.0 & 36.0 & 41.0 & 38.0 & 57.0 & 59.0 & 28.0 & 41.0 & 3.0 & 26.0 & 37.5 \\
    DeepSeek-VL2-Tiny & 32.0 & 52.0 & 36.0 & 24.0 & 27.0 & 25.0 & 47.0 & 27.0 &26.0 & 4.0 & 16.0 & 28.7 \\
    Pixtral-12B & 31.0 & \cellcolor{gray!30}68.0 & 24.0 & 24.0 & 21.0 & 38.0 & 53.0 & 21.0 & 24.0 & 3.0 & 37.0 & 31.3 \\
    LLaVA-OneVision-7B & 42.0 & 59.0 & 21.0 & 41.0 & 30.0 &  50.0 & 59.0 & 26.0 & 20.0 & 0.0 & 22.0 & 33.6 \\
    EMU3 & 31.0 & 52.0 & 24.0 & 25.0 & 17.0 & 25.0 & 47.0 & 25.0 & 24.0 & 0.0 & 20.0 & 26.4 \\
    \shline
    \rowcolor{yellow!15}\multicolumn{13}{l}{\textit{\textbf{Baseline}}} \\
    Random Guessing & 33.0 & 50.0 & 25.0 & 25.0 & 20.0 & 25.0 & 50.0 & 25.0 & 25.0 & 4.2 & 20.0 & 27.5 \\
    $\uparrow$ Random ($p < 0.05$) & 42.0 & 59.0 & 33.0 & 33.0 & 28.0 & 33.0 & 59.0 & 33.0 & 33.0 & 9.0 & 28.0 & 35.5 \\
    \shline
    \end{tabular}}
    
    \label{tab:main_results}
\end{table*}

\begin{table*}[!t]
    \centering
    \caption{\small\textbf{Comparing top-performing MLLMs with human proficiency on LEGO-Puzzles-Lite.} The best results are marked in \textbf{bold}. The top 3 overall performances are highlighted in \colorbox{green}{\small Dark Green}, \colorbox{green!30}{\small Medium Green}, and \colorbox{green!10}{\small Light Green}, respectively.}
    \vspace{-3mm}
    \resizebox{\textwidth}{!}{%
    \tablestyle{3pt}{1.3}
    \begin{tabular}{lccccccccccc|r}
    \shline
    \multirow{2}{*}{\textbf{Models}} & \multicolumn{4}{c}{\cellcolor{cyan!5}\textbf{Spatial Understanding}} & \multicolumn{4}{c}{\cellcolor{orange!10}\textbf{Single-Step Reasoning}} & \multicolumn{3}{c|}{\cellcolor{brown!15}\textbf{Multi-Step Reasoning}} & \multirow{2}{*}{\textbf{Overall}}\\
    & Height & Adjacency & Rotation & Multiview & Next-Step & Dependency & Rotation Stat. & Position & Backwards & Ordering & Outlier \\
    \shline
    \rowcolor{yellow!15}\multicolumn{13}{l}{\textit{\textbf{LEGO-Puzzles-Lite}}} \\
    \textbf{Human proficiency} & \textbf{70.0} & \textbf{95.0} & \textbf{95.0} & \textbf{100.0} & \textbf{90.0} & \textbf{100.0} & \textbf{100.0} & \textbf{95.0} & \textbf{95.0} & \textbf{95.0} & \textbf{95.0} & \cellcolor{green} 93.6 \\
    GPT-5 & 55.0 & 70.0 & 80.0 & 85.0 & 75.0 & 95.0 & 55.0 & 60.0 & 70.0 & 85.0 & 60.0 & \cellcolor{green!30} 71.8 \\
    GPT-o3 & 65.0 & 70.0 & 75.0 & 70.0 & 75.0 & 95.0 & 55.0 & 55.0 & 80.0 & 75.0 & 65.0 & \cellcolor{green!10}70.9 \\
    Gemini-2.5-Pro & 40.0 & 70.0 & 70.0 & 55.0 & 85.0 & 90.0 & 50.0 & 50.0 & 55.0 & 80.0 & 80.0 &  65.9 \\
    Qwen3-VL-32B-Instruct & 45.0 & 55.0 & 30.0 & 50.0 & 70.0 & 95.0 & 65.0 & 75.0 & 80.0 & 50.0 & 50.0 & 60.5 \\
    InternVL3-78B  & 50.0 & 65.0 & 35.0 & 50.0 & 65.0 & 75.0 & 60.0 & 40.0 & 20.0 & 30.0 & 40.0 & 48.2 \\
    Qwen2.5-VL-72B & 25.0 & 70.0 & 25.0 & 35.0 & 65.0 & 70.0 & 65.0 & 45.0 & 55.0 & 20.0 & 55.0 & 48.2 \\
    \bottomrule
    \end{tabular}}
    \label{tab:human_results}
\end{table*}


\begin{figure*}[!t]
    \centering
    \includegraphics[width=\linewidth]{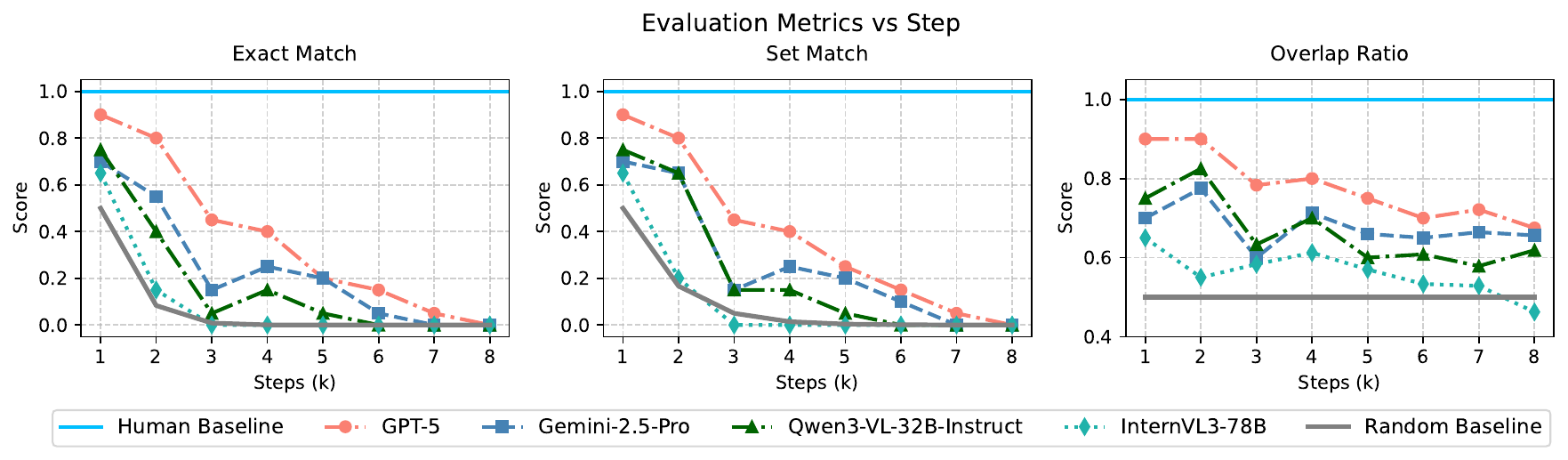}
    \caption{\small{\textbf{Evaluation on \textit{PLAN-$k$-Step}} of different MLLMs across varying numbers of steps.}}
    \vspace{-4mm}
    \label{fig:plan-k-step}
\end{figure*}

\subsection{Results}

\label{sec:main_results}
We report results on the Elementary Set in \cref{tab:main_results} and \cref{tab:human_results}. 
We summarize key findings as below.

\vspace{-2mm}
\paragraph{Clear Gap Between Proprietary and Open-Source Models.}
As shown in \cref{tab:main_results}, there is a clear performance gap between proprietary and open-source MLLMs. GPT-5 exhibits strong spatial reasoning capabilities, achieving overall accuracies of 72.0\%, while most open-source MLLMs perform only marginally better than $\uparrow$\textit{Random}.

\vspace{-2mm}
\paragraph{Human Outperforms the Best MLLMs by Over 20\%.}
To study the performance gap between human and MLLMs on the Elementary set of LEGO-Puzzles, 
we randomly select 20 questions from each task to create \textbf{LEGO-Puzzles-Lite}, resulting in a total of 220 QA pairs, 
and use this subset for investigation. 
We invited 30 human experts to solve these questions, ensuring a diverse and representative evaluation of human-level reasoning. 
As shown in \cref{tab:human_results}, human experts consistently achieve significantly higher overall performance (93.6\%). In contrast, current MLLMs fall short, with even the most advanced models, GPT-5 and GPT-o3, trailing over 20\% behind human performance across all tasks.

\vspace{-2mm}
\paragraph{Hard Cases in the Elementary Set.}
As shown in \cref{tab:main_results}, most models (21/29) perform worse than $\uparrow$\textit{Random} in the \textit{Height} task.
Part of the cause is that we observe that MLLMs tend to answer questions based on the object's height in its 2D projection rather than in its true 3D perspective. See appendix
for some examples.
This is consistent with the observation in 3DSRBench~\cite{ma20243dsrbench}, where MLLMs exhibit a similar tendency on real-world images.
Our construction of the \textit{Height} task contains many images that can lead to different answers between 2D and 3D perspectives, which further exacerbates the problem.

Another task that MLLMs struggle with is the \textit{Rotation} task, where 16 out of 29 models fall below $\uparrow$\textit{Random}.
This shows that current MLLMs may not perceive and distinguish object orientation changes reliably, which could 

Beyond basic spatial understanding, increasing the reasoning complexity further degrades the performance of MLLMs. For example, when the task changes from \textit{Next-Step} to \textit{Ordering}, the accuracy drops significantly for most models.

\vspace{-2mm}
\paragraph{MLLMs Perform Poor in Multi-Step Reasoning.}
LEGO-Puzzles also reveals substantial limitations in MLLMs' sequential reasoning capabilities, particularly when multiple reasoning steps are required.
As shown in \cref{tab:main_results}, almost half of the models perform below \textit{$\uparrow$ Random} in the \textit{Ordering} task, with some models (e.g., InternVL2.5-8B, LLaVA-OneVision-7B) failing completely. Similar trends are observed in the \textit{Backwards} task, where 12 out of 20 open-source models perform below \textit{$\uparrow$ Random}.
To further explore the upper bound of MLLMs in multi-step reasoning, we conduct the more challenging and length-controllable task in~\cref{sec:planningset}.


\renewcommand{\thefootnote}{\fnsymbol{footnote}}

\section{Experiment on the Planning Set}

\label{sec:planningset}
\subsection{VQA Results}

\label{subsec:nextkstep} 
To further investigate the long-horizon multi-step sequential reasoning abilities of MLLMs, we design a scalable and challenging task \textbf{\textit{PLAN-$k$-Step-VQA}}, which controls the number of reasoning steps $k$ to model the realistic reasoning planning and execution across multiple sequences under noisy conditions (defined in \cref{sec:task-Planning Set}).
We select the four top-performing models in \cref{sec:ex}: GPT-5, Gemini-2.5-Pro, Qwen3-VL-32B-Instruct, and InternVL3-78B.

\vspace{-2mm}
\paragraph{Evaluation Metrics.}
We use \textit{exact match accuracy} as in the Elementary set, and additionally include two metrics: (1) \textit{set match accuracy}(\%), defined as the proportion of predictions in which the predicted steps contain exactly the same elements as the ground truth, regardless of order. (2) \textit{overlap ratio} (\%), which measures the fraction of predicted steps that also appear in the ground-truth sequence without considering order.
The results are shown in \cref{fig:plan-k-step}.

\vspace{-2mm}
\paragraph{Performance Degradation when $k$ Increases.} 
As shown in \cref{fig:plan-k-step}, all models exhibit a clear decline in \textit{exact match accuracy} as the number of planning steps $k$ increases. GPT-5 drops from $90\%$ at $k=1$ to $0\%$ at $k=8$, while both Qwen3-VL-32B-Instruct and InternVL3-78B reach $0\%$ once $k>5$. For $k>2$, the accuracy of all models drops below 50\%. 

\vspace{-2mm}
\paragraph{Model Struggles in Selecting the Correct Set of Steps. }
Comparing the accuracy of \textit{exact match} and \textit{set match}, GPT-5 exhibits almost identical curves. 
This indicates that errors emerge already at the set-selection stage: the model fails to identify the correct elements and consequently selects an incorrect set of steps, let alone ordering them correctly.

\vspace{-2mm}
\paragraph{Clear Gap Between Open-Source and Proprietary Models.}  
For \textit{overlap ratio}, even when \textit{exact match accuracy} falls to 0\% at $k=7/8$, the proprietary models still achieve $\sim$65\%, indicating that they often identify a substantial portion of the correct steps. Conversely, the open-source models degrade to near random-guessing levels as $k>4$.

\vspace{-2mm}
\paragraph{MLLMs Perform Far Below Human Level.} 
To establish a human baseline, we invited five expert participants to solve the same set of tasks. 
Humans achieved a perfect score of 100\% across all values of $k$, showing no decline as the number of steps increased. 
On average, they completed all tasks in about 85 minutes. 
This demonstrates that, unlike MLLMs, humans can reliably handle long-horizon multi-step reasoning tasks. 
In contrast, current MLLMs struggle to track and integrate spatial transformations across multiple steps, where accumulated errors lead to inconsistent predictions in longer reasoning chains.

%% file: 5_generation.tex
\subsection{Image Generation Results}

\label{sec:generation}

Experiments in \cref{subsec:nextkstep} show that MLLMs struggle to maintain consistent multi-step reasoning as the number of steps increases, even in a simplified multiple-choice setting. To further investigate their generative ability, we construct a full-stage planning task (defined in \cref{sec:task-Planning Set}) that requires models to generate intermediate assembly states step by step.

\begin{figure}[!t]
    \centering
    \includegraphics[width=\linewidth]{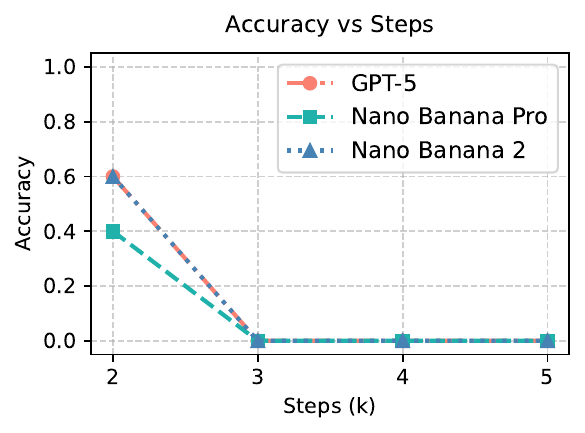}
    \vspace{-7mm}
    \caption{
        \textbf{Quantitative results of generative full-stage planning.}
        The figure shows the quantitative results for GPT-5, Nano Banana Pro and Nano Banana 2 across scaling steps $k=2,3,4,5$.}
    \vspace{-3mm}
    \label{fig:generation_fullstage_acc}
\end{figure}

\begin{figure*}[!t]
\centering
\includegraphics[width=\linewidth]{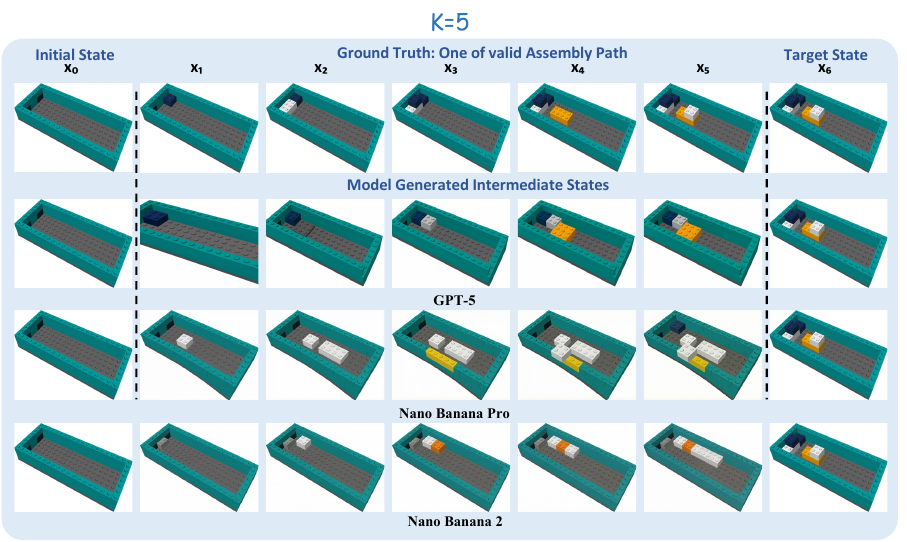}
\caption{
    \textbf{Qualitative results of generative full-stage planning.}
    The figure provides qualitative examples of the generated intermediate states for both models at $k=5$.
}
\vspace{-3mm}
\label{fig:generation_fullstage}
\end{figure*}

\vspace{-3mm}
\paragraph{Benchmark Models.}
We evaluate three strong current generation models: GPT-5,
Gemini-3-pro-image-preview (Nano Banana Pro) and gemini-3.1-flash-image-preview (Nano Banana 2). 

\vspace{-3mm}
\paragraph{Evaluation Protocol.}
To construct ground-truth planning paths, we first apply topology expansion to enumerate all valid assembly sequences from the initial state to the final state under physical and assembly constraints, 
where lower parts must be assembled before upper parts. 
This produces multiple feasible topological trajectories rather than a single fixed path. 
We then perform multi-turn generation, where the model outputs one intermediate state at each turn conditioned on the previously generated history.

For evaluation, each generated step is matched against all topology-expanded candidates at that step using mean squared error (MSE).
After per-step matching, we check whether the matched steps can be concatenated into a complete valid sequence in the topology-expanded set. 
If the matched steps form a full legal trajectory, the case is counted as correct (score 1); otherwise it is counted as incorrect (score 0).
This automatic evaluation process makes it possible to quantitatively assess the correctness of the generated multi-step planning paths, which would be infeasible to evaluate manually due to the large number of possible trajectories.

We set the planning step $k = 2, 3, 4, 5$. 
We do not evaluate longer horizons because even at $k=3$ the strongest models already fail completely. 
Although some generated images show reasonable appearance similarity, none of the models succeed in producing a fully correct generative planning trajectory. 
We construct 50 cases for this experiment, with 10 cases for each planning horizon $k$.
The quantitative results and qualitative examples are shown in \cref{fig:generation_fullstage_acc} and \cref{fig:generation_fullstage}, respectively.

\vspace{-4mm}
\paragraph{Even the strongest current MLLMs fail to perform full-stage generative planning.}
The examples in \cref{fig:generation_fullstage} show that the models cannot produce reasonable intermediate states, with errors often appearing from the very first step. 
This suggests that current models still lack the ability to perform consistent multi-step reasoning and apply such reasoning to generate correct intermediate images. 
True generative multi-step planning requires not only spatial understanding, but also consistent visual imagination, temporal reasoning, and implicit physical constraints, which remain challenging for current MLLMs. More analysis is provided in \cref{sec:dis}.

%% file: 6_discussion.tex
\section{Discussion}
\label{sec:dis}

\subsection{Correlation with Real-world Spatial Reasoning}

\begin{table}[!t]
    \centering
    \caption{Pearson correlation coefficients (PCC) and p-values for \textit{height} and \textit{adjacency} tasks.}
    \begin{tabular}{lcc}
        \hline
        \textbf{Task} & \textbf{PCC} & \textbf{P-value} \\
        \hline
        Height & 0.93 & 0.00723 \\
        Adjacency & 0.98 & 0.00046 \\
        \hline
    \end{tabular}
    \label{tab:correlation_results}
\end{table}
While LEGO-Puzzles is built on rendered data, it aims to evaluate fundamental spatial reasoning capabilities that are also essential in real-world scenarios. To assess its generalizability beyond synthetic environments, we compare model performance on LEGO-Puzzles with 3DSRBench~\cite{ma20243dsrbench}, a benchmark based on natural images. Both datasets contain conceptually similar tasks—specifically, the \textit{Height} task in LEGO-Puzzles aligns with \textit{Height} in 3DSRBench, and \textit{Adjacency} in LEGO-Puzzles corresponds to the \textit{Location} task in 3DSRBench.

We evaluate all proprietary models tested in LEGO-Puzzles on the corresponding tasks in 3DSRBench and compute the Pearson correlation coefficient~\cite{benesty2009pearson} to measure consistency in performance across the two datasets. As shown in \cref{tab:correlation_results}, the results reveal strong positive correlations: 0.93 for \textit{Height} and 0.98 for \textit{Adjacency}, both statistically significant ($p < 0.01$).

These findings suggest that LEGO-Puzzles not only offers high scalability and precise control in synthetic settings but also captures spatial reasoning patterns that generalize well to natural images. This validates its utility as a proxy for evaluating real-world spatial understanding.

\subsection{MLLMs Struggle with Elementary Tasks \\in Image Generation}
\label{subsec:generation_elementary}

The results in \cref{sec:generation} show that even at the very first step, the generated images often contain errors, such as misaligned parts, incorrect orientations, or missing pieces. To better understand the source of these failures, we conduct an additional experiment where the model is required to generate only a single turn. 
We select five tasks from the \textit{Elementary} set of LEGO-Puzzles: \textit{Rotation*}, \textit{Multiview*}, \textit{Position*}, \textit{Dependency*}, and \textit{Next-Step*}. For each task, we sample 20 questions, resulting in a total of 100 questions. 
In this experiment, we change the output format of the tasks from multiple choice to image generation. Instead of selecting from predefined options, the model is required to directly generate the visual output.

\begin{figure*}[!t]
\vspace{-3mm}
\centering
\includegraphics[width=\linewidth]{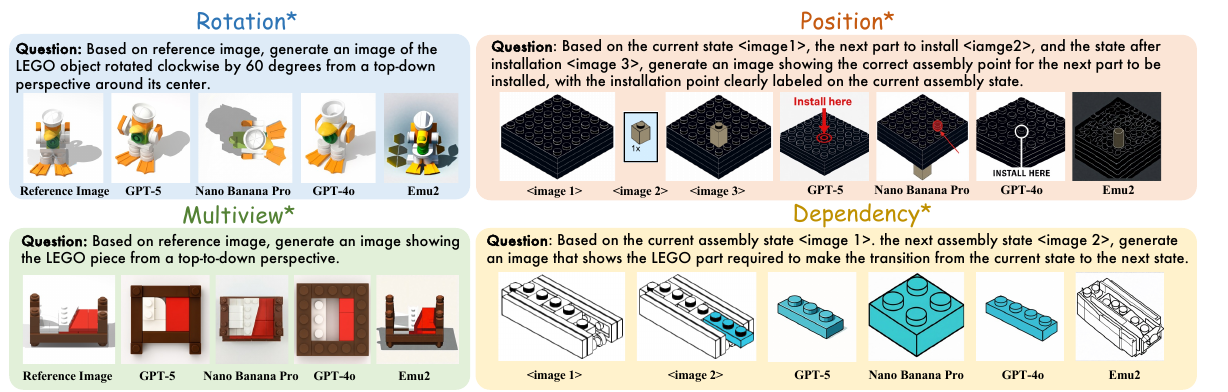}
\caption{\textbf{Qualitative visual results for image generation tasks.} Note: The questions above are slightly simplified for clarity and brevity.}
\label{fig:generation}
\vspace{-3mm}
\end{figure*}

\vspace{-2mm}
\paragraph{Models and Evaluation.}
We evaluate the open-source models Emu2~\cite{sun2023emu2chat}, GILL~\cite{koh2023generating}, and Anole~\cite{chern2024anole}, as well as the proprietary models GPT-5, GPT-4o,
Gemini-3-pro-image-preview (Nano Banana Pro) and Gemini-2.0-Flash, all of which support long-range sequence input and image output. 
For evaluation, traditional metrics such as FID~\cite{heusel2017gans}, CLIPScore~\cite{hessel2021clipscore}, and X-IQE~\cite{chen2023x} mainly assess image fidelity or cross-modal alignment, often relying on pre-trained model priors or fixed scoring heuristics. They struggle to capture the fine-grained spatial accuracy required in LEGO assembly tasks, where even small errors such as misaligned parts or incorrect orientations can invalidate the result. Furthermore, many recent multimodal evaluation metrics depend on GPT-based models~\cite{liu2024holistic}, introducing uncontrollable bias into the evaluation process. Therefore, we enlist 5 human experts to assess model performance across two dimensions: appearance similarity and instruction following. Each aspect is rated on a scale from 0 to 3. Detailed scoring guidelines are provided in the appendix.

\begin{table*}[!t]
    \vspace{2mm}
    \centering
    \caption{\textbf{Evaluation on \textit{generation}.} We conduct human-based evaluation to assess the ``Appearance'' (App) and ``Instruction Following'' (IF) scores of GPT-5, GPT-4o, Nano Banana Pro, Gemini-2.0-Flash, Emu2, GILL, and Anole, using a scoring scale from 0 to 3 for both dimensions.}
    \vspace{-3mm}
    \resizebox{\linewidth}{!}{%
    \tablestyle{8pt}{1.3}
    \begin{tabular}{l|cc|cc|cc|cc|cc|cc|cc}
        \shline
        \multirow{2}{*}{\textbf{Task \textbackslash MLLM}} & \multicolumn{2}{c}{\textbf{GPT-5}} & \multicolumn{2}{c}{\textbf{Nano Banana Pro}} & 
        \multicolumn{2}{c}{\textbf{GPT-4o}} & \multicolumn{2}{c}{\textbf{Gemini-2.0-Flash}} &  \multicolumn{2}{c}{\textbf{Emu2}} & \multicolumn{2}{c}{\textbf{GILL}} & \multicolumn{2}{c}{\textbf{Anole}} \\ \scline{2-15}
        & App & IF & App & IF & App & IF & App & IF & App & IF & App & IF & App & IF \\ 
        \shline
        Rotation* & 2.45 & 1.75& 2.60 & 1.70 & 2.35 & 1.75 & 2.05 & 1.45 &  1.70 &0.00 & 0.00 & 0.00 & 0.10 & 0.00 \\
        Multiview* & 1.90 & 2.10& 2.45 & 2.55 & 2.30 & 2.00 & 2.40 & 1.65 & 1.65 & 0.25 & 0.00 & 0.00 & 0.05 & 0.00 \\
        Position* & 2.75 & 1.60& 3.00 & 2.30 & 2.30 & 1.65 & 2.85 & 1.30 & 0.50 & 0.00 & 0.00 & 0.00 & 0.00 & 0.00 \\
        Dependency* & 1.85 & 2.20 & 2.05 & 1.65 & 2.05 & 2.00 & 1.70 & 0.95 & 0.55 & 0.00 & 0.00 & 0.00 & 0.00 & 0.00 \\
        Next-Step* & 2.30 & 1.75 & 2.50 & 1.75 & 2.25 & 1.45 & 1.75 & 0.05 & 0.05 & 0.00 & 0.00 & 0.00 & 0.00 & 0.00 \\
        \shline
        Overall & 2.25 & 1.88 & 2.52 & 1.99 & 2.25 & 1.77 & 2.15 & 1.08 & 0.89 & 0.05 & 0.00 & 0.00 & 0.03 & 0.00 \\
        \shline
    \end{tabular}}
    \label{tab:generation}
    \vspace{-2mm}
\end{table*}

\vspace{-2mm}
\paragraph{MLLMs struggle to perform single-turn image generation, not alone full-stage multi-step planning.}
\cref{tab:generation} presents the human evaluation results across five generation tasks. Overall, proprietary models outperform open-source ones in both appearance consistency (App) and instruction adherence (IF). Among them, 
Nano Banana Pro achieves the highest overall scores (App: 2.52, IF: 1.99), followed by GPT-5 (App: 2.25, IF: 1.88).
However, both models show clear room for improvement in instruction following. 
GPT-4o and Gemini-2.0-Flash perform slightly worse than GPT-5 and Nano Banana Pro, with lower scores in both dimensions, especially in instruction following (IF: 1.77 and 1.08, respectively).
Among open-source models, Emu2 shows some ability to preserve visual appearance (App: 0.89) but fails almost entirely in instruction following (IF: 0.05), treating the task more as image replication than reasoning-based generation. 
GILL and Anole perform the worst, with near-zero scores across all tasks and frequently irrelevant outputs. The qualitative results are shown in \cref{fig:generation}. More cases are provided in the appendix.

Overall, these results show that current models, especially open-source ones, still struggle with spatial reasoning-based image generation in LEGO assembly tasks. Even the strongest models, while producing outputs with high appearance consistency, often fail to fully adhere to the instructions, indicating a gap between visual understanding and precise generation.

%% file: 7_conclusion.tex
\section{Conclusion}
\label{sec:con}

We introduce LEGO-Puzzles, a novel benchmark specifically designed to evaluate spatial understanding, as well as single-step and multi-step sequential reasoning in MLLMs. 
Inspired by human cognitive patterns in LEGO construction, we create two task sets. 
The Elementary set includes over 1,100 carefully curated visual question-answering (VQA) samples across 11 distinct tasks, providing diverse scenarios to assess multimodal visual reasoning. 
The Planning set directly requires the model to generate a step-by-step plan for assembling a target LEGO structure in VQA and image generation settings.
We conduct comprehensive experiments with 29 advanced MLLMs, revealing substantial performance gaps compared to humans, particularly in long-horizon planning and the generation of spatially coherent visual outputs. 
These findings underscore the urgent need to enhance the spatial understanding and sequential reasoning capabilities of multimodal AI.

%% file: 8_appendix.tex
\clearpage
\refstepcounter{section}
\section*{Appendix}
\addcontentsline{toc}{section}{Appendix}
\setcounter{subsection}{0}
\renewcommand{\thesubsection}{\Alph{subsection}}
\renewcommand{\thesubsubsection}{\thesubsection.\arabic{subsubsection}}



\subsection{Error Cases in \textit{Height} Task}
\label{app:height_error}
\begin{figure}[h]
\centering
\vspace{-3mm}
\includegraphics[width=1\linewidth]{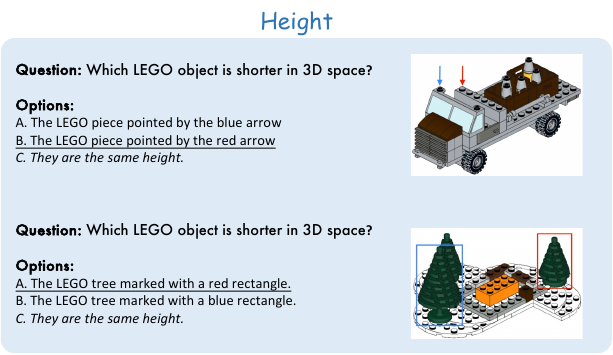}
\caption{
\small{\textbf{Error cases in the Height task}. \underline{Correct answers are underlined}, and \textit{human-selected answers are shown in italics.}
\textit{Note: Questions are slightly simplified for clarity and brevity.}}}
\label{fig:height_error}
\vspace{-3mm}
\end{figure}

The \textit{Height} task in the Elementary set contains intentional 2D–3D ambiguities that can lead to occasional human mistakes. In these cases, humans may rely on quick visual intuition rather than careful spatial reasoning, especially when the blocks are not adjacent and the height must be inferred through spatial imagination and reasoning rather than direct comparison. When objects are farther apart or partially occluded, people may overlook subtle depth differences and misjudge the true height. We also found that these mistakes largely come from not inspecting the images carefully; when participants were explicitly reminded to pay closer attention, they achieved 100\% accuracy. Several such cases are illustrated in \cref{fig:height_error}.

\subsection{Problem statistics in the Elementary Set of LEGO-Puzzles}

\begin{figure}[h]
\centering
\includegraphics[width=0.5\linewidth]{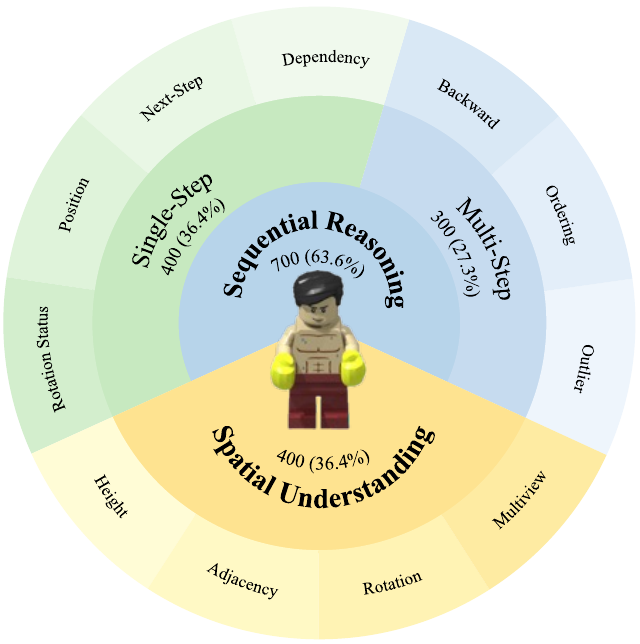}
\caption{\textbf{Problem statistics in the Elementary Set of LEGO-Puzzles}}
\label{fig:data_structure}
\end{figure}

As illustrated in \cref{fig:data_structure}, the Elementary Set contains three levels.

\noindent\textbf{\textit{Level 1:} Spatial Understanding. }
(1) \textit{Height}
(2) \textit{Adjacency}
(3) \textit{Rotation}
(4) \textit{Multiview}

\noindent\textbf{\textit{Level 2:} Single-Step Sequential Reasoning. }
(5) \textit{Rotation Status}
(6) \textit{Position}
(7) \textit{Next-Step}
(8) \textit{Dependency}

\noindent\textbf{\textit{Level 3:} Multi-Step Sequential Reasoning. }
(9) \textit{Backwards}
(10) \textit{Ordering}
(11) \textit{Outlier}

The Elementary set consists of over $1,100$ visual question-answering (VQA) pairs derived from 407 LEGO building instructions, encompassing $11$ tasks across spatial understanding, single-step and multi-step sequential reasoning. Each task contains 100 samples to ensure balanced evaluation across categories.



\subsection{Evaluation}
\label{app:evaluation}
In this section, we provide the set of prompt templates used in our evaluation. All prompts match the requirements of each task type and ensure consistent formatting, minimal ambiguity, and controlled instruction scope. Below, we organize the prompts by task group.
\subsubsection{Prompts used for LEGO-Puzzles Evaluation}
\label{app:prompt_lego}
This section lists the prompt templates used in the Elementary Set of LEGO-Puzzles benchmark. Each template corresponds to one of the eleven core tasks defined in ~\cref{sec:task-Elementary Set} of the main paper, including \textit{Spatial Understanding}, \textit{Single-step Sequential Reasoning}, and \textit{Multi-step Sequential Reasoning}. For each task, the prompt defines the model's role, describes the input structure (including image tokens), and specifies the required response format.

\begin{promptbox}[Prompt for Task \textit{Height}]

You are a specialized LEGO 3D height analyzer. Your primary task is to compare the heights of LEGO objects based on their positions in 3D space. You will be provided with an image containing 3D LEGO objects. Your answers should be based solely on the provided LEGO 3D data, without any additional assumptions. Keep your responses clear, direct, and focused on the question. Please respond with only the letter corresponding to your choice (A, B, or C).\promptnewline\promptnewline LEGO object is shorter in 3D space?\promptnewline\promptnewline Here is the 3D LEGO scene: \imagetoken{1} Options: A. The LEGO tree marked with a red rectangle. B. The LEGO tree marked with a blue rectangle. C. They are the same height. Please select the correct answer from the options above.\promptnewline

\end{promptbox}

\begin{promptbox}[Prompt for Task \textit{Adjacency}]

You are a specialized LEGO 3D adjacency analyzer. You will be provided with an image containing 3D LEGO objects. Your goal is to determine whether two LEGO objects are directly touching (adjoining) or not (separated) based on their positions in 3D space. Your answers should be based solely on the provided LEGO 3D data, without any additional assumptions. Keep your responses clear, direct, and focused on the question. Please respond with only the letter corresponding to your choice (A or B).\promptnewline\promptnewline Are the LEGO trees pointed to by the two red arrows adjoining or separated?\promptnewline\promptnewline Here is the 3D LEGO scene: \imagetoken{1} Options: A. Adjoining. B. Separated. Please select the correct answer from the options above.\promptnewline

\end{promptbox}

\begin{promptbox}[Prompt for Task \textit{Multiview}]

You are a specialized LEGO 3D multi-view analyzer. Your primary task is to identify the correct perspective of LEGO object images in a 3D scene. You will be provided with one reference image (x\_0) and four optional images showing the LEGO object from different viewpoints. Your goal is to determine which option corresponds to one of the following perspectives: top-down, left-to-right, right-to-left, or front-to-back. Your answers should be based solely on the provided LEGO 3D data, without any additional assumptions. Keep your responses clear, direct, and focused on the question. Please respond with only the letter corresponding to your choice (A, B, C, or D).\promptnewline\promptnewline Based on the LEGO object shown in the reference image (x\_0), which of the following images shows the LEGO object from a front-to-back perspective?\promptnewline\promptnewline Reference image (x\_0): \imagetoken{1} Options: A. \imagetoken{2} B. \imagetoken{3} C. \imagetoken{4} D. \imagetoken{5} Please select the correct answer from the options above.\promptnewline

\end{promptbox}

\begin{promptbox}[Prompt for Task \textit{Rotation}]

You are a specialized LEGO 3D rotation analyzer. You will be provided with two images: Image 1 and Image 2. Image 2 shows the same LEGO object as Image 1, but rotated clockwise around its center from a top-down perspective (looking down on the LEGO object from above) by one of the following angles: $30^\circ$, $60^\circ$, $90^\circ$, or $120^\circ$. Your task is to determine how many degrees the LEGO object in Image 2 has rotated clockwise relative to Image 1. Your answers should be based solely on the provided LEGO 3D data, without any additional assumptions. Keep your responses clear, direct, and focused on the question. Please respond with only the letter corresponding to your choice (A, B, C, or D).\promptnewline\promptnewline How many degrees has the LEGO object in Image 2 rotated clockwise around its center relative to Image 1?\promptnewline\promptnewline Image 1 (original): \imagetoken{1} Image 2 (rotated): \imagetoken{2} Options: A. $30^\circ$ B. $60^\circ$ C. $90^\circ$ D. $120^\circ$ Please select the correct answer from the options above.\promptnewline

\end{promptbox}

\begin{promptbox}[Prompt for Task \textit{Dependency}]

You are a specialized LEGO 3D assembly analyzer. Your primary task is to identify the correct LEGO piece needed to transition from one assembly state to another. You will be provided with two consecutive assembly state images: the current assembly state (x\_1) and the next assembly state (x\_2), followed by four LEGO piece images. Your goal is to select the correct option that shows the LEGO piece required to make the transition from the first state to the second. Your answers should be based solely on the provided sequence of images, focusing on the logical relationships between the sequence of images and the 3D spatial relationships between the LEGO pieces. Avoid making any additional assumptions beyond the information conveyed in the sequence. Keep your responses clear, direct, and focused on the question. Please respond with only the letter corresponding to your choice (A, B, C, or D).\promptnewline\promptnewline This is the current assembly state (x\_1): \imagetoken{1} This is the next assembly state (x\_2): \imagetoken{2}\promptnewline\promptnewline Please choose which option correctly shows the LEGO piece needed for the transition: Options: A. \imagetoken{3} B. \imagetoken{4} C. \imagetoken{5} D. \imagetoken{6} Please select the correct answer from the options above.\promptnewline

\end{promptbox}

\begin{promptbox}[Prompt for Task \textit{Next-Step}]

You are a LEGO 3D next step prediction analyzer. Your primary task is to identify the correct next state of a LEGO object assembly. You will be provided with the current assembly state (x\_1), the next LEGO piece to be added (x\_2), and the target object image (x\_3), which represents the final completed object. Your goal is to select the correct option that shows the assembly state after adding the next LEGO piece. The next LEGO piece (x\_2) is a step toward achieving the final object (x\_3), and your task is to determine how adding this LEGO piece could potentially move the assembly toward the final object. Your answers should be based solely on the provided sequence of images, focusing on the logical relationships between the sequence of images and the 3D spatial relationships between the parts. Avoid making any additional assumptions beyond the information conveyed in the sequence. Keep your responses clear, direct, and focused on the question. Please respond with only the letter corresponding to your choice (A, B, C, D, or E).\promptnewline\promptnewline This is the current assembly state (x\_1): \imagetoken{1} This is the next LEGO piece to be used (x\_2): \imagetoken{2} This is the target object (x\_3): \imagetoken{3} Please choose which option correctly shows the state of the object after adding the specified LEGO piece: Options: A. \imagetoken{4} B. \imagetoken{5} C. \imagetoken{6} D. \imagetoken{7} E. \imagetoken{8} Please select the correct answer from the options above.\promptnewline

\end{promptbox}

\begin{promptbox}[Prompt for Task \textit{Rotation Status}]

You are a specialized LEGO 3D rotation and assembly analyzer. You will be provided with three images: the base structure (x\_0), the final structure (x\_1), and the additional piece (x\_2). Your task is to analyze the images and decide whether rotating the additional piece (x\_2) is necessary to attach it to the base structure (x\_0) and form the final structure (x\_1). Your answers should be based solely on the provided LEGO 3D data, without any additional assumptions. Keep your responses clear, direct, and focused on the question. Please respond with only the letter corresponding to your choice (A: ``Yes'' or B: ``No'').\promptnewline\promptnewline Does the additional piece (x\_2) need to be rotated to attach it to the base structure (x\_0) in order to form the final structure (x\_1)? Base structure (x\_0): \imagetoken{1} Final structure (x\_1): \imagetoken{2} Additional piece (x\_2): \imagetoken{3} Options: A. Yes. B. No. Please select the correct answer from the options above.\promptnewline

\end{promptbox}

\begin{promptbox}[Prompt for Task \textit{Position}]

You are a LEGO 3D assembly position analyzer. Your primary task is to determine the correct assembly point of a given LEGO piece based on the current state and the next state. You will be provided with images representing the current state (x\_0), the LEGO piece to install (x\_1), the state after installation (x\_2), and installation options. Your goal is to analyze the given images and determine which of the four options shows the correct assembly point for the next LEGO piece. Your answers should be based solely on the provided LEGO 3D data, without any additional assumptions. Keep your responses clear, direct, and focused on the question. Please respond with only the letter corresponding to your choice (A, B, C, or D).\promptnewline\promptnewline Based on the current state (x\_0), the next LEGO piece (x\_1) to install, and the state after installation (x\_2), which of the following images shows the correct installation point? Current state (x\_0): \imagetoken{1} Part to install (x\_1): \imagetoken{2} State after installation (x\_2): \imagetoken{3} Options: A. \imagetoken{4} B. \imagetoken{5} C. \imagetoken{6} D. \imagetoken{7} Please select the correct answer from the options above.\promptnewline

\end{promptbox}

\begin{promptbox}[Prompt for Task \textit{Ordering}]

You are a specialized LEGO 3D step ordering analyzer. Your primary task is to order the correct sequence of LEGO assembly steps to build a target model. You will be provided with the current assembly state image (x\_1), the target assembly state image (x\_2), and four step images. Your goal is to arrange the four step images in the correct order (e.g., ``BDAC'') that transitions from the current state to the target state. Your answers should be based solely on the provided sequence of images, focusing on the logical relationships between the sequence of images and the 3D spatial relationships between the parts. Avoid making any additional assumptions beyond the information conveyed in the sequence. Keep your responses clear, direct, and focused on the question. Please respond with only the sequence of letters corresponding to your choice (e.g., ``BDAC'').\promptnewline\promptnewline This is the current assembly state (x\_1): \imagetoken{1}\promptnewline\promptnewline This is the target assembly state (x\_2): \imagetoken{2}\promptnewline\promptnewline Please arrange the following options (A, B, C, and D) in the correct order to transition from the current state to the target state: A. \imagetoken{3} B. \imagetoken{4} C. \imagetoken{5} D. \imagetoken{6} Please select the correct answer from the options above.\promptnewline

\end{promptbox}

\begin{promptbox}[Prompt for Task \textit{Outlier}]

You are a specialized LEGO 3D step sequence analyzer. You will be provided with: the current assembly state image (x\_1) and the target assembly state image (x\_2), followed by five step images. Four of these steps correctly fit in the sequence needed to transition from x\_1 to x\_2, while one does NOT belong in this sequence. Your task is to identify the one incorrect step. Your answers should be based solely on the provided sequence of images, focusing on the logical relationships between the sequence of images and the 3D spatial relationships between the parts. Avoid making any additional assumptions beyond the information conveyed in the sequence. Keep your responses clear, direct, and focused on the question. Please respond with only the letter corresponding to your choice (A, B, C, D, or E).\promptnewline\promptnewline This is the current assembly state (x\_1): \imagetoken{1} This is the target assembly state (x\_2): \imagetoken{2}\promptnewline\promptnewline Please choose which option does NOT belong in this sequence: Options: A. \imagetoken{3} B. \imagetoken{4} C. \imagetoken{5} D. \imagetoken{6} E. \imagetoken{7} Please select the correct answer from the options above.\promptnewline

\end{promptbox}

\begin{promptbox}[Prompt for Task \textit{Backwards}]

You are a specialized LEGO 3D reverse engineering analyzer. You will be provided with the final target assembly image and four step images. Your goal is to select the correct option that represents the step to assemble the target model. The incorrect options may be rejected for reasons such as: incorrect LEGO piece colors, incorrect LEGO piece positions, incorrect LEGO piece orientations, incorrect number of LEGO pieces, etc. Your answers should be based solely on the provided sequence of images, focusing on the logical relationships between the sequence of images and the 3D spatial relationships between the LEGO pieces. Avoid making any additional assumptions beyond the information conveyed in the sequence. Keep your responses clear, direct, and focused on the question. Please respond with only the letter corresponding to your choice (A, B, C, or D).\promptnewline\promptnewline This is the target assembly image: \imagetoken{1} Which option correctly shows the correct assembly step? Options: A. \imagetoken{2} B. \imagetoken{3} C. \imagetoken{4} D. \imagetoken{5} Please select the correct answer from the options above.\promptnewline

\end{promptbox}

 
\subsubsection{Human Evaluation in Image Generation}

\label{app:human_generation}
\paragraph{Evaluation protocol.}
To evaluate the quality of model-generated images (\cref{subsec:generation_elementary} of the main paper), we use two complementary dimensions: \textit{Appearance Similarity (App)} and \textit{Instruction Following (IF)}. Each dimension is scored on a scale from 0 to 3, where higher scores reflect closer alignment with the expected visual outcome. Five expert annotators independently score each sample, and final scores are averaged across annotators. Below, we describe the evaluation criteria in detail.

\vspace{-2mm}

\paragraph{Dimension 1: Appearance Similarity (App).}
This score measures how visually similar the generated image is to the expected ground-truth output, focusing on global and local visual elements such as object shape, part layout, color, and scene background. It does not consider whether the model followed instructions—only how much the output visually resembles the intended result.
\begin{description}[leftmargin=12mm,labelwidth=10mm,labelsep=2mm,itemsep=1pt,topsep=2pt]
    \item[\textbf{Score 3}] \textbf{Nearly identical.} All major and minor visual details match, apart from imperceptible or stylistic differences.
    \item[\textbf{Score 2}] \textbf{Mostly similar.} Most key features are correct, with only minor visual discrepancies (e.g., slightly incorrect positioning or color of one part).
    \item[\textbf{Score 1}] \textbf{Partially similar.} Some features are correct, but major visual elements are incorrect or missing.
    \item[\textbf{Score 0}] \textbf{Dissimilar.} The overall shape, structure, or color is incorrect and does not meaningfully match the target.
\end{description}

\vspace{-2mm}
\paragraph{Dimension 2: Instruction Following (IF).}
IF measures whether the generated image performs the transformation requested by the prompt. Unlike App, it focuses on task correctness rather than general visual similarity. Because the required transformation differs across tasks, we use the following task-specific rubrics.

\begin{taskrubric}{\textit{Rotation*}}
\begin{description}[leftmargin=12mm,labelwidth=10mm,labelsep=2mm,itemsep=1pt,topsep=2pt]
    \item[\textbf{Score 3}] The object is rotated in the correct direction and by the correct angle.
    \item[\textbf{Score 2}] The object is rotated in the correct direction but deviates slightly from the target angle.
    \item[\textbf{Score 1}] The object is rotated in the wrong direction, or the rotation magnitude is clearly incorrect.
    \item[\textbf{Score 0}] No rotation is applied, or the output is irrelevant.
\end{description}
\end{taskrubric}

\begin{taskrubric}{\textit{Multiview*}}
\begin{description}[leftmargin=12mm,labelwidth=10mm,labelsep=2mm,itemsep=1pt,topsep=2pt]
    \item[\textbf{Score 3}] The viewpoint precisely matches the target perspective.
    \item[\textbf{Score 2}] The viewpoint changes in the correct direction but with slight deviation from the target view.
    \item[\textbf{Score 1}] The viewpoint changes in the wrong direction or is clearly inaccurate.
    \item[\textbf{Score 0}] The viewpoint is unchanged, or the output is irrelevant.
\end{description}
\end{taskrubric}

\begin{taskrubric}{\textit{Dependency*}}
\begin{description}[leftmargin=12mm,labelwidth=10mm,labelsep=2mm,itemsep=1pt,topsep=2pt]
    \item[\textbf{Score 3}] The correct part is generated with accurate shape and color.
    \item[\textbf{Score 2}] The part is mostly correct, with minor errors in color or minor structural additions.
    \item[\textbf{Score 1}] The generated part is noticeably different in shape or color from ground truth.
    \item[\textbf{Score 0}] The generated part is irrelevant, or the instruction is not followed.
\end{description}
\end{taskrubric}

\begin{taskrubric}{\textit{Position*}}
\begin{description}[leftmargin=12mm,labelwidth=10mm,labelsep=2mm,itemsep=1pt,topsep=2pt]
    \item[\textbf{Score 3}] The part is placed at the precise target location.
    \item[\textbf{Score 2}] The placement is close to the target but slightly offset.
    \item[\textbf{Score 1}] The placement is far from the target location.
    \item[\textbf{Score 0}] No attempt to identify a placement, or the output is irrelevant.
\end{description}
\end{taskrubric}

\begin{taskrubric}{\textit{Next-Step*}}
\begin{description}[leftmargin=12mm,labelwidth=10mm,labelsep=2mm,itemsep=1pt,topsep=2pt]
    \item[\textbf{Score 3}] The output correctly shows the structure after the new part is added.
    \item[\textbf{Score 2}] The output reflects the instruction generally but contains moderate deviations.
    \item[\textbf{Score 1}] The output contains only superficial or incorrect modifications.
    \item[\textbf{Score 0}] The output ignores instruction or is unrelated to the task.
\end{description}
\end{taskrubric}

\paragraph{Annotator calibration.}
Before evaluation, all annotators are trained using a set of labeled examples. Disagreements were resolved via discussion.

\subsection{Data Curation}
\noindent\textbf{Question–Answer Generation.}
\label{app:qa_gen}
To support scalable and structured benchmark construction, we develop a fully template-based QA generation pipeline tailored to all tasks in LEGO-Puzzles.

\textit{Question Template.}
For each task, we manually design a task-specific question format that defines the model's role, introduces the input images through an explicit token such as \imagetoken{x}, and specifies the expected output form.
This ensures that every QA pair adheres to a unified structure while preserving the distinct requirements of each task. Each template consists of:
\vspace{1mm}
\begin{itemize}
\item \textbf{Task-specific instruction.} Defines the model's expected expertise (e.g., ``You are a specialized LEGO 3D rotation analyzer.''), ensuring consistent reasoning behavior across samples within the same task.
\item \textbf{Image-referencing token.} All visual inputs are introduced through explicit tokens such as \imagetoken{x}, providing a standardized way to embed image information across tasks.
\item \textbf{Task-specific question format.} Specifies the exact objective for the task.
\end{itemize}
\vspace{1mm}
Appendix~\ref{app:prompt_lego} lists the full set of Question templates used in the Elementary set of LEGO-Puzzles.

\textit{Ground-Truth Answer.}
Ground-truth answers are generated automatically from the structured assembly data (for sequential reasoning tasks) or from human annotation (for spatial understanding tasks).
For Single-Step and Multi-Step Sequential Reasoning tasks, every LEGO project is constructed step-by-step with precise metadata (e.g., part ID, assembly-step index). Thus, the ground-truth answer for each question can be automatically extracted.
For example, in the \textit{Next-Step task}, given the current assembly-step index $k$ and the final target assembly-step index $m$, the ground-truth next state is simply the image corresponding to assembly-step index $k + 1$. This provides a clear and deterministic answer. For Spatial Understanding tasks, each task involves geometric or perceptual judgments that cannot be derived solely from metadata. Their ground-truth answers are therefore obtained through human annotation to ensure correctness and consistency.

\subsection{LLM-as-a-judge in Image Generation Evaluation}
To explore the potential of LLMs as automatic evaluators for image generation tasks in Section 5.2 of the main paper, we conduct a systematic comparison between human expert scores and LLM-based (GPT-4o) automatic evaluation on image outputs generated by some
representative models in our benchmark (Gemini-2.0-Flash, GPT-4o, GILL and Anole) across two evaluation dimensions: Appearance Similarity (App) and Instruction Following (IF).

\noindent\textbf{We implemented a structured LLM-based evaluation pipeline.}
Each generated sample was evaluated by GPT-4o using a detailed rubric mirroring our human annotation protocol. 
The prompt included the original image(s), the instruction, the model-generated image, and the ground-truth image. 
GPT-4o assigned 0--3 scores for Appearance Similarity and Instruction Following, following the same criteria used by human experts. 
We then compared GPT-4o's scores with human judgments using Pearson correlation, MAE, and MSE.

\noindent\textbf{LLM-as-a-judge is not yet suitable to serve as a reliable judge in this setting and exhibits low consistency with human evaluation on strong models.}
As shown in \cref{tab:llm_judge_strong}, GPT-4o demonstrates substantial disagreement with human ratings when evaluating high-quality generations from Gemini-2.0-Flash and GPT-4o. 
For example, Pearson correlation on Appearance Similarity is only 0.133 for Gemini-2.0-Flash and even negative (-0.443) for GPT-4o. 
MAE and MSE are also large: GPT-4o reaches 0.82 MAE / 0.716 MSE on Appearance Similarity, while Gemini-2.0-Flash shows 1.80 MAE / 3.56 MSE. 
Deviations remain notable on Instruction Following as well (e.g., 0.77 MAE and 0.762 MSE for Gemini-2.0-Flash). 
These results suggest that current MLLMs still lack robust spatial understanding of structures and are therefore \textbf{not yet suitable to serve as reliable judges} for spatially grounded image-generation tasks that require precise spatial alignment.

\begin{table}[h]
\centering
\small
\caption{Consistency between GPT-4o (LLM-as-a-judge) and human evaluation on outputs from strong models. We report Pearson correlation, MAE, and MSE for Appearance Similarity (App) and Instruction Following (IF). Lower MAE/MSE and higher Pearson indicate better agreement with human judgments.}
\label{tab:llm_judge_strong}
\begin{tabular}{lcc}
\toprule
\textbf{Model} & \textbf{GPT-4o (APP / IF)} & \textbf{Gemini-2.0-Flash (APP / IF)} \\
\midrule
Pearson & -0.443 / 0.298 & 0.133 / 0.685 \\
MAE & 0.820 / 0.430 & 1.800 / 0.770 \\
MSE & 0.716 / 0.333 & 3.560 / 0.762 \\
\bottomrule
\end{tabular}
\end{table}
\noindent\textbf{LLM-as-a-judge aligns with human evaluation only on clearly failed cases.}
We further evaluated the worst-performing models in our benchmark (GILL and Anole), whose generations are largely unrelated to the input images and instructions. 
As shown in \cref{tab:llm_judge_weak} , both human annotators and GPT-4o assigned near-zero scores across both dimensions, indicating that existing MLLMs can reliably identify severe failures but struggle with finer-grained judgments.

\noindent\textbf{In conclusion}, while we agree that reproducible automatic evaluation is desirable, our analysis shows that current LLM-as-a-judge approaches do not yet replicate human judgment on spatially precise, instruction-grounded image-generation tasks.
Human evaluation therefore remains essential. Improving spatially aware LLM-based evaluators is an important direction for future work.

\subsection{Cases of Plan-k-Step-Generation}
\begin{figure*}[!t]
\centering
\includegraphics[width=\textwidth]{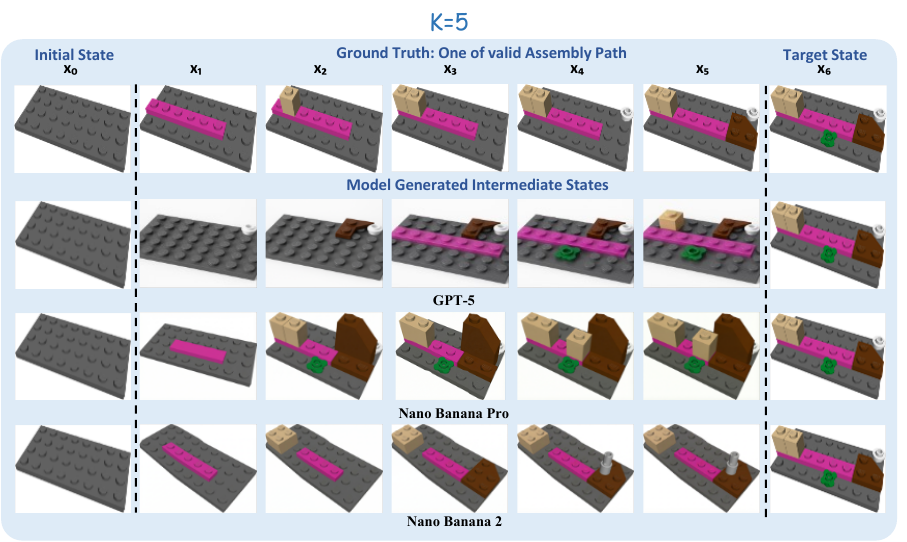}
\caption{
    \textbf{Qualitative results of generative full-stage planning.}
    The figure provides qualitative examples of the generated intermediate states for both models at $k=5$.
}
\label{fig:generation_fullstage2}
\end{figure*}

\begin{figure*}[!t]
\centering
\includegraphics[width=\textwidth]{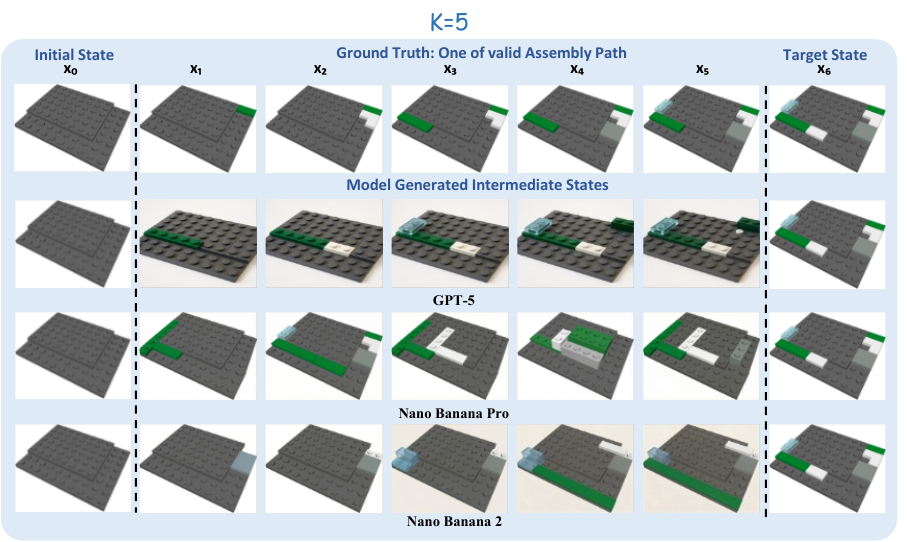}
\caption{
    \textbf{Qualitative results of generative full-stage planning.}
    The figure provides qualitative examples of the generated intermediate states for both models at $k=5$.
}
\label{fig:generation_fullstage3}
\end{figure*}

To further illustrate the capabilities and limitations of MLLMs in generative planning tasks, we provide additional qualitative examples of the Plan-k-Step-Generation task (\cref{sec:planningset} of the main paper).
As illustrated in \cref{fig:generation_fullstage2,fig:generation_fullstage3}, these cases highlight the limitation of current MLLMs in generating accurate intermediate assembly states, especially as the number of steps increases.

\subsection{Cases of the Elementary Tasks in Image Generation}

\label{app:generation_case}
To better illustrate model behavior in spatial reasoning through image generation, we provide more qualitative examples of three representative tasks: \textit{Rotation*}, \textit{Multiview*}, and \textit{Dependency*} (\cref{subsec:generation_elementary} of the main paper). These cases highlight common success and failure modes across models, revealing how different MLLMs handle viewpoint changes, object rotations, and part-level reasoning. In each case, the task setup and the generated responses are visualized, with questions simplified for clarity. Examples are selected to reflect typical model behavior observed during evaluation.

\begin{figure*}[!t]
\centering
\includegraphics[width=\textwidth]{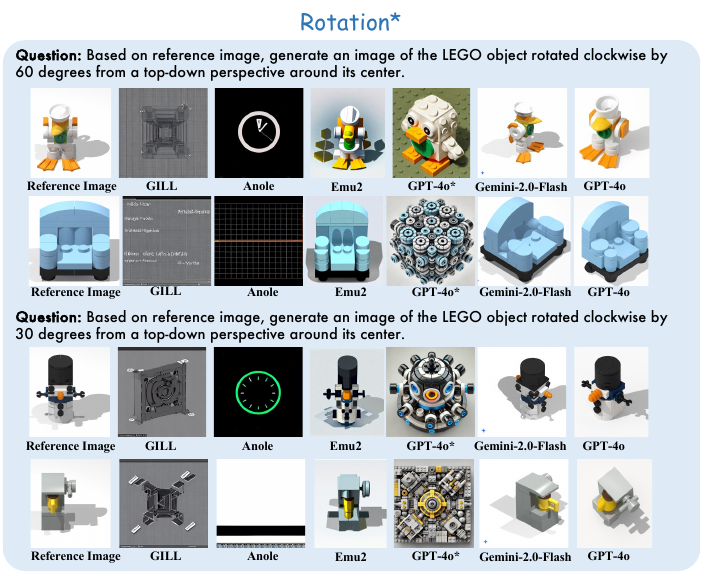}
\vspace{-5mm}
\caption{Qualitative visual generation results for Task \textit{Rotation*}. Note: The questions above are slightly simplified for clarity and brevity. GPT-4o* denotes the version of GPT-4o released prior to March 6, 2025.}
\label{fig:appendix_generation_rotation}
\end{figure*}

\begin{figure*}[!t]
\centering
\includegraphics[width=\textwidth]{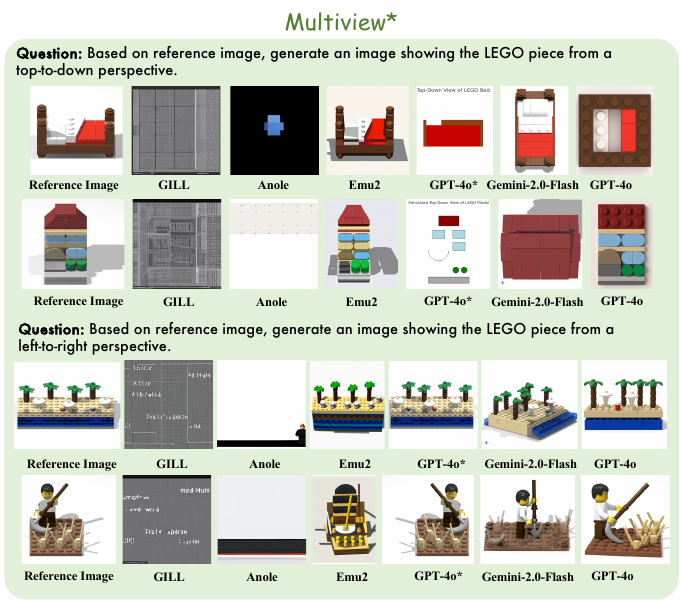}
\vspace{-5mm}
\caption{Qualitative visual generation results for Task \textit{Multiview*}. Note: The questions above are slightly simplified for clarity and brevity. GPT-4o* denotes the version of GPT-4o released prior to March 6, 2025.}
\label{fig:appendix_generation_multiview}
\end{figure*}

\begin{figure*}[!t]
\centering
\includegraphics[width=\textwidth]{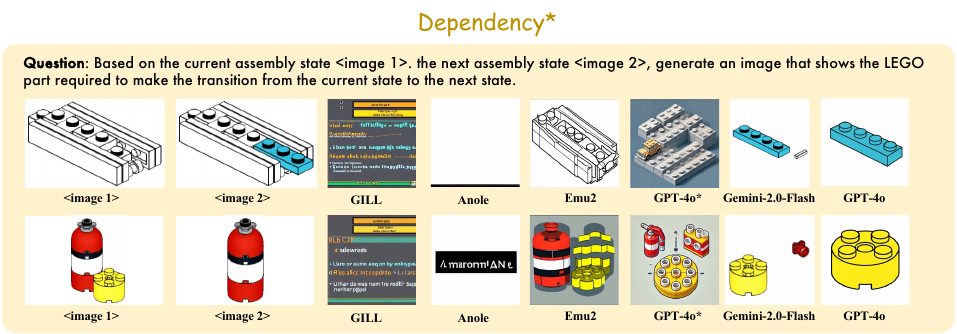}
\vspace{-5mm}
\caption{Qualitative visual generation results for Task \textit{Dependency*}. Note: The questions above are slightly simplified for clarity and brevity. GPT-4o* denotes the version of GPT-4o released prior to March 6, 2025.}
\label{fig:appendix_generation_dependency}
\end{figure*}

\cref{fig:appendix_generation_rotation} shows examples for the \textit{Rotation*} task, where models are asked to generate rotated versions of a LEGO object.
\cref{fig:appendix_generation_multiview} presents \textit{Multiview*} generation cases, illustrating how models interpret viewpoint transformations. \cref{fig:appendix_generation_dependency} displays \textit{Dependency*} cases, which assess whether models can generate the correct part needed for a given assembly transition.

\begin{table*}[!t]
\centering
\small
\caption{Comparison between human scores and GPT-4o evaluation on clearly failed generations from weak models (GILL and Anole). Both human annotators and GPT-4o assign near-zero scores across tasks, indicating that LLM-as-a-judge can identify severe failures but struggles with finer-grained judgments.}
\label{tab:llm_judge_weak}
\resizebox{\textwidth}{!}{
\begin{tabular}{lcccc}
\toprule
\textbf{Task} & \textbf{GILL Human (APP / IF)} & \textbf{GILL GPT-4o (APP / IF)} & \textbf{Anole Human (APP / IF)} & \textbf{Anole GPT-4o (APP / IF)} \\
\midrule
rotation & 0.00 / 0.00 & 0.00 / 0.00 & 0.10 / 0.00 & 0.15 / 0.05 \\
multi & 0.00 / 0.00 & 0.05 / 0.05 & 0.05 / 0.00 & 0.10 / 0.10 \\
position & 0.00 / 0.00 & 0.00 / 0.00 & 0.00 / 0.00 & 0.00 / 0.00 \\
dep & 0.00 / 0.00 & 0.00 / 0.00 & 0.00 / 0.00 & 0.00 / 0.00 \\
next & 0.00 / 0.00 & 0.00 / 0.00 & 0.00 / 0.00 & 0.00 / 0.00 \\
\midrule
Average & 0.00 / 0.00 & 0.01 / 0.01 & 0.03 / 0.00 & 0.05 / 0.03 \\
\bottomrule
\end{tabular}
}
\end{table*}

